\documentclass[preprint,11pt,a4paper,authoryear,nonatbib]{elsarticle}

\usepackage[
left=2.54cm,
right=2.54cm,
top=2.54cm,
bottom=2.54cm
]{geometry}

\usepackage{setspace}
\usepackage{amsmath, amsthm, amssymb, appendix, bm, graphicx, hyperref, mathrsfs, booktabs, multirow, colonequals, xcolor}

\newtheorem{theorem}{Theorem}
\newtheorem{corollary}[theorem]{Corollary}

\newtheorem{assumption}{Assumption}

\usepackage[T1]{fontenc}    
\usepackage{url}            
\usepackage{amsfonts}       
\usepackage{nicefrac}       
\usepackage{float}

\usepackage{leftidx}
\usepackage{fancyhdr}

\usepackage{tabularx}
\usepackage[backend=biber,style=authoryear,natbib=true,sorting=nyt,giveninits=true,maxcitenames=2,uniquename=false,uniquelist=false]{biblatex}
\usepackage{caption}
\usepackage{longtable}
\usepackage{array}
\usepackage{tikz}

\newcommand{\ff}{\textup{FFN}}

\newcommand{\dhd}{d_{\mathrm{embd}}}

\newcommand{\tdepth}{L_{T}}
\newcommand{\ffdepth}{L_\ff}
\newcommand{\ffwidth}{w_\ff}
\newcommand{\clen}{l}

\newcommand{\token}{h}

\def\block(#1,#2)#3{\multicolumn{#2}{c}{\multirow{#1}{*}{$ #3 $}}}

\usepackage{lmodern}
\usepackage{microtype}
\usepackage{xr}
\usepackage{xurl}
\usepackage{etoolbox}
\journal{Journal of Econometrics}
\hypersetup{hidelinks}
\makeatletter
\let\ps@pprintTitle\ps@plain
\makeatother
\allowdisplaybreaks[1]

\begin{document}
	
	\begin{frontmatter}
		\title{Conditional Asset Pricing Transformer}
		\author[aff1]{Hao Fu}
		\author[aff1]{Xinbing Kong\corref{cor1}}
		\author[aff1]{Zeyu Li}
		\author[aff1]{Junfan Mao}
		\cortext[cor1]{Corresponding author. Email: xinbingkong@126.com.}
		\affiliation[aff1]{
            organization={Southeast University},
            addressline={Jiulonghu Campus, No. 2 Southeast University Road, Jiangning District},
            city={Nanjing},
            state={Jiangsu},
            postcode={211189},
            country={P.R. China}
        }
		\begin{abstract}
			In this paper, we propose a conditional asset pricing transformer (CAP-T) that captures nonlinear interactions among firms through attention to forecast excess returns. In the first stage, a transformer estimates conditional expected returns, and kernel-weighted principal component analysis of its residuals estimates the local factor loading space. Projecting the conditional mean onto this space separates factor risk compensation from conditional mispricing. Then, in the second stage, separate transformers learn these two components, whose predictions are combined to forecast returns. As a result, the proposed framework decomposes returns into conditional mispricing, factor risk compensation, a centered factor component, and an unpredictable innovation. We establish approximation and generalization bounds and characterize how first-stage prediction errors propagate through local factor estimation and component learning, providing statistical guaranties for conditional mispricing, factor loadings, and factor risk premia. Finally, simulations and empirical studies demonstrate the predictive and interpretive benefits of the proposed method.
		\end{abstract}
        
		\begin{keyword}
        Asset pricing;
        Transformers;
        Cross-firm attention;
        Temporal dependence.
        \end{keyword}
	\end{frontmatter}

    \section{Introduction}
    \label{sec:introduction}
    
    Understanding differences in expected returns is a central task
    of asset pricing. It informs the allocation of capital, the
    measurement of investment performance, and the assessment of
    compensation for risk. The foundations laid by
    \citet{Sharpe1964} and \citet{Ross1976} connect expected returns
    to systematic exposures, placing the relationship between an
    individual asset and the broader market at the center of the
    pricing problem. As increasingly rich information about firms
    becomes available, a central challenge is to determine how that
    information enters expected returns and what it reveals about
    their economic sources.

Factor models provide a parsimonious framework for addressing
this challenge. They summarize common variation in a large
collection of returns through a small number of factors, making
the estimation of systematic exposures feasible in large panels
\citep{BaiNg2002}. Their usefulness for pricing depends on both
the information retained by the estimated factors and the
evolution of the loading space. \citet{LettauPelger2020} show
that factors important for expected returns can be difficult to
recover from return variation alone and incorporate pricing
information into factor estimation. \citet{SuWang2017} develop
local PCA for models with smoothly varying loadings, while \citet{kong2017number,kong2018systematic} independently introduce high-frequency local PCA with four types of general varying loadings. Robustness to heavy tails has also motivated Huber-loss-based
estimators for vector and tensor factor models
\citep{BarigozziHeLiTrapani2026,HeLiLiuZhou2025}.
These developments illustrate how factor estimation can
accommodate features of financial data while preserving an
interpretable representation of common risk.

With the inclusion of observable firm characteristics, we can use conditional factor models to connect these characteristics to systematic risk exposures. Remarkably, \citet{IPCA} develop a novel
instrumented principal component analysis (IPCA) framework, allowing
characteristics to determine factor loadings through a linear
mapping. To deal with non-linearity, \citet{SuJinWang2025} develop sieve estimation for state-varying
factor models, allowing loadings to be unknown individual-specific
functions of observed state variables. Recently, deep learning has opened new possibilities for capturing richer non-linear relationships between characteristics and returns. For example, \citet{GuKellyXiu2020} document the predictive value of
nonlinear methods for equity returns, while
\citet{GuKellyXiu2021} use conditional autoencoders to model
nonlinear dependence of factor exposures on characteristics.
A notable advance in this direction is the structural
framework of \citet{Fan1}, which brings flexible prediction
and economic interpretation together. Their DNN-varying
approach combines neural network estimation at each date
with local PCA to separate conditional mispricing from risk
compensation and subsequently learn predictive functions
for both components. This framework gives nonlinear return
prediction a clear economic structure, distinguishing
the sources of expected returns while allowing each
component to be learned separately.

However, the characteristic mappings in these pricing
models rely primarily on each firm's own information.
This leaves an important dimension of the pricing problem
less fully explored: firms operate within a shared economic
environment and are connected through production,
competition, and financing. The information contained in
a firm's profitability or investment may therefore become
more informative when viewed alongside the characteristics
of other firms. Such connections have observable pricing
implications, for instance, \citet{CohenFrazzini2008} document return
predictability along customer and supplier relationships,
highlighting the relevance of information originating
outside the firm itself. To exploit structured cross-information in
high-dimensional data, \citet{WangLiuChen2019} develop
matrix factor models that preserve the row and column
structure of matrix-valued time series. Then, projection-based methods further improve loading-space
estimation in matrix and tensor factor models
\citep{YuHeKongZhang2022,HanChenYangZhang2024,
BarigozziHeLiTrapani2026}.
Related work develops CP-decomposition-based representations
for matrix time series \citep{ChangHeYangYao2023}
and tests for the presence and dimensions of row and column
factor spaces \citep{HeKongTrapaniYu2023}. Meanwhile, \citet{ChenLiLiLinton2025} estimate time-varying Granger-causality
and partial-correlation networks for high-dimensional time series,
with extensions that account for common factors. Together, these studies motivate pricing models that
use firm characteristics jointly and allow the relevance
of other firms' information to vary with economic conditions.
    
    Transformers offer another conceptually natural way to model these cross-firm relationships.
    Through attention, they combine information across firms
    using weights that depend on the observed characteristics
    \citep{Attentionisallyouneed,LeeEtAl2019}.
    In asset pricing, \citet{KellyKuznetsovMalamudXu2025}
    incorporate attention into stochastic discount factor
    estimation to share information across assets. In parallel, our focus is on conditional expected returns and on
    how information across firms contributes differently
    to risk compensation and conditional mispricing. To the best of our knowledge, this is the first transformer framework for conditional asset pricing that jointly attends to a broad cross section of assets to learn risk compensation and conditional mispricing separately.
    Indeed, predicting expected returns alone does not distinguish compensation for systematic risk from conditional mispricing, and the distinction between these components is central to our approach. Note that the relationships among firms that inform risk compensation may differ from those that inform conditional mispricing, and learning only their sum leaves these differences implicit. Allowing each component to draw on
    different aspects of the characteristic information serves an interpretive purpose as well as an predictive one.

    Finally, establishing statistical guaranties for non-linear asset pricing models often requires
    connecting flexible function approximation with estimation
    under dependence. Econometric work on sieve estimation and subsequent estimation
stages addresses weak dependence and the effects of estimated
functions \citep{ChenShen1998,ChenLiao2015}, whereas \citet{Wang2022GeneralizedFactors} studies inference in
factor-augmented regressions and forecasts when latent factors
are estimated from nonlinear or mixed data. For neural networks, \citet{SchmidtHieber2020} derive
    regression rates that exploit compositional structure. Then, \citet{YunEtAl2020} establish universal
    approximation results for transformers, and \citet{manifold} develop
    approximation and statistical learning theory on manifolds
    of low intrinsic dimension. Finally, in the context of conditional asset pricing, \citet{Fan1} establish statistical
guaranties linking neural network estimation to factor decomposition.

    In this work, we develop a Conditional Asset Pricing Transformer, denoted
    CAP-T, that integrates these features within a joint
    estimation framework. The paper makes four contributions.

    First, we broaden the information used in conditional
    asset pricing from individual firm characteristics to
    the joint characteristics of the market cross section.
    Compared with the firm-specific pricing functions in
    \citet{Fan1}, CAP-T allows other firms' characteristics
    to inform an individual firm's expected return.
    Attention learns these relationships from the observed
    characteristics without requiring a prespecified network
    of economic links. This formulation accommodates
    relationships that change with market conditions and
    allows different firms to provide relevant information
    for different pricing components.
    
    Second, inspired by the integration of economic structure
    and nonlinear learning in \citet{GuKellyXiu2021,Fan1},
    we use an identified factor decomposition to organize
    learning from information across firms. Predicting
    expected returns alone leaves the sources of
    predictability unresolved: the same characteristic
    may convey information about risk compensation,
    conditional mispricing, or both. Our decomposition
    provides separate targets for these economic components,
    allowing their networks to learn different relationships
    among firms before combining their predictions.
    The value of this structure is therefore twofold:
    it gives the predictions an economic interpretation
    and permits each component to use the joint information
    according to its own predictive relationships.
    
    Third, we establish approximation and statistical error
    bounds for this estimation procedure. Building on the
    manifold approximation analysis of \citet{manifold},
    we develop a construction for pricing functions with
    vector outputs whose approximation complexity is
    governed by the intrinsic dimension of the joint
    characteristic state. Our statistical analysis allows
    geometric $\beta$-mixing and changing marginal state
    distributions. It also addresses the dependence between
    successive estimation stages, a central concern in
    \citet{ChenLiao2015}, by tracing errors from the initial
    regression through local PCA and component learning. For fixed dimensions, the initial regression
    achieves a squared $L^2$ error rate of
    $T^{-2\beta/(2\beta+d)}$ up to logarithmic factors,
    where $\beta$ is smoothness and $d$ is the intrinsic dimension.
    The component estimators and their sum inherit this rate
    with an additional loading estimation error term.
    
    Fourth, we document predictive gains alongside economic
    structure in the learned representations. Complementing
    the evidence on nonlinear return prediction in
    \citet{GuKellyXiu2020}, our simulations examine recovery
    of the underlying pricing components, while the equity
    application shows improved return prediction relative
    to IPCA and DNN-varying and useful portfolio signals
    after transaction costs. With five factors and semiannual updating,
    the three layer CAP-T (ReLU) mispricing portfolio achieves
    a net annualized Sharpe ratio of $0.85$, compared with
    at most $0.69$ for DNN-varying and $0.67$ for IPCA. Interestingly, the attention representations are able to
    cluster firms along distinct economic dimensions:
    risk representations emphasize volatility and trading
    conditions, whereas alpha representations distinguish
    profitability and financing constraints. Moreover, selected firm
    associations also find support in corporate disclosures
    and characteristic similarity. The analysis on characteristic importance further 
    provides complementary evidence, with risk prediction
    emphasizing market information and alpha placing greater
    weight on valuation and investment.  These findings connect the model's predictive performance to interpretable
    differences in the information used for risk compensation
    and conditional mispricing.
    
    The remainder of the paper is organized as follows.
    Section~\ref{sec:model} presents the model and estimation
    procedure. Section~\ref{sec:theory} develops the theoretical
    results. Section~\ref{simulation} reports the simulation
    evidence, and Section~\ref{real data} presents the empirical
    analysis. Section~\ref{sec:conclusion} gives further discussion.

	\section{The Model and Methodology}
	\label{sec:model}
    In this section, we develop CAP-T within a conditional factor framework that separates expected excess returns into risk compensation and conditional mispricing. We first specify the model and its identifying restrictions. Then we introduce the transformer pricing functions that incorporate information across firms. In the end, we present an estimation procedure that uses local factor estimation to construct the targets for learning the two pricing components.

    \subsection{Conditional Factor Model}

    Inspired by \citet{Fan1}, we consider a conditional factor model that separates risk compensation from conditional mispricing. Let $\bm y_t\in\mathbb R^N$ denote excess returns for $N$ firms at date $t$, and let $X_{t-1}\in\mathbb R^{N\times P}$ collect their $P$ characteristics observed before these returns are realized. We specify
    \begin{equation}
        \label{eq:1}
        \bm y_t
        =g_\alpha(X_{t-1})
        +\bm\beta(X_{t-1})\bm\lambda(X_{t-1})
        +\bm\beta(X_{t-1})\bm\zeta_t+\bm u_t,
    \end{equation}
    where $g_\alpha(X_{t-1})\in\mathbb R^N$ denotes conditional mispricing, $\bm\beta(X_{t-1})\in\mathbb R^{N\times K}$ contains exposures to $K$ latent factors, and $\bm\lambda(X_{t-1})\in\mathbb R^K$ denotes factor risk premia. The vectors $\bm\zeta_t\in\mathbb R^K$ and $\bm u_t\in\mathbb R^N$ are factor innovations and idiosyncratic shocks, respectively. We assume that the two innovation sequences are independent and identically distributed over time, mutually independent, and centered. Current innovations are independent of past information, implying
    \begin{equation}
        \label{eq:conditional_zero_mean}
        \mathbb E(\bm\zeta_t\mid X_{t-1})=\bm 0,
        \quad
        \mathbb E(\bm u_t\mid X_{t-1})=\bm 0.
    \end{equation}

    Define risk compensation as $g_{\mathrm{risk}}(X):=\bm\beta(X)\bm\lambda(X)$. Taking conditional expectations in \eqref{eq:1} under \eqref{eq:conditional_zero_mean} decomposes expected excess returns into the sum of two components:
    \begin{equation}
        \label{eq:3}
        f(X_{t-1}):=\mathbb E(\bm y_t\mid X_{t-1})
        =g_\alpha(X_{t-1})+g_{\mathrm{risk}}(X_{t-1}).
    \end{equation}
    Equation~\eqref{eq:3} expresses expected excess returns as the sum of conditional mispricing and risk compensation. Our statistical objective is to recover and separately learn these two components. To identify the two components, we assume that the loading matrix has full column rank and impose
    \begin{equation}
        \label{eq:alpha_loading_orthogonality}
        \bm\beta(X)^\top g_\alpha(X)=\bm 0.
    \end{equation}
    Under \eqref{eq:alpha_loading_orthogonality}, the decomposition in \eqref{eq:3} identifies risk compensation as the projection onto the loading space and mispricing as the orthogonal remainder. Factor risk premia are the corresponding coefficients in the chosen loading coordinates.
    
    Following the structural estimation strategy of \citet{Fan1},
    we proceed in three steps. First, we estimate the conditional
    mean $f$. Second, we apply local PCA to the resulting
    residuals and use the estimated loading space to construct
    targets for the two pricing components. Third, we learn each
    component separately and combine their predictions to
    forecast excess returns.
	
	\subsection{Transformer Pricing Functions}
	
	The proposed conditional asset pricing transformer maps the joint characteristic state
	$X\in\mathbb R^{N\times P}$ to an $N$-dimensional vector of
	conditional return components. That is, we seek a compact representation of firm characteristics,
	using attention to combine predictive information across firms.
	Attention allows the pricing functions to incorporate interactions among firms \citep{Attentionisallyouneed,LeeEtAl2019,KellyKuznetsovMalamudXu2025}. For a single attention head, let
	$W_Q,W_K,W_V\in\mathbb R^{P\times d^{'}}$ denote the learned
	query, key, and value weight matrices. The attention
	operation is
	\begin{equation}
		\label{eq:firm_attention}
		S(X)
		=
		\sigma\!\left(
		\frac{(XW_Q)(XW_K)^\top}{\sqrt{d^{'}}}
		\right),
		\quad
		\mathcal A(X)=S(X)XW_V,
	\end{equation}
	where $\sigma$ denotes the attention activation,
	$S(X)\in\mathbb R^{N\times N}$, and
	$\mathcal A(X)\in\mathbb R^{N\times d^{'}}$.
	The matrix $(XW_Q)(XW_K)^\top$ collects cross products of
	the projected firm characteristics. Its $(i,j)$th entry
	measures the interaction between the query representation
	of firm $i$ and the key representation of firm $j$.
	After activation, $S_{ij}(X)$ determines the contribution
	of firm $j$ to the updated representation of firm $i$. The resulting pricing functions can therefore use
	characteristics throughout the cross section, with the strength of these interactions varying with the observed state. Quite often, multiple heads based on \eqref{eq:firm_attention} capture
    different patterns of interaction. Their outputs are
    concatenated and projected to form
    \begin{equation*}
        \label{eq:multihead_pricing}
        H(X)
        =
        \operatorname{concat}
        \bigl(\mathcal A_1(X),\ldots,\mathcal A_m(X)\bigr)W_O,
        \quad
        W_O\in\mathbb R^{md'\times d'},
    \end{equation*}
    where each head uses separate weight matrices.
    A shared feedforward transformation is applied to each
    firm's representation. Writing this representation as
    a column vector $\token\in\mathbb R^{d'}$, we have
    \begin{equation*}
        \label{eq:feedforward_pricing}
        \ff(\token)
        =
        W_{\ffdepth}
        \sigma\left(
        W_{\ffdepth-1}\dots
        \sigma\left(W_1\token+b_1\right)
        \dots+b_{\ffdepth-1}
        \right)
        +b_{\ffdepth},
    \end{equation*}
    where $\ffdepth$ denotes the feedforward depth and
    the weight matrices and biases have compatible dimensions.
    Attention and feedforward transformations are combined
    through residual connections and stacked across layers.
    An output map then produces the $N$ conditional return
    components.
    
    Let $\mathcal F$ denote the class of transformer maps
    $F_\theta:\mathbb R^{N\times P}\to\mathbb R^N$
    over which the pricing functions are estimated, where $\theta$ varies across its members\footnote{The transformer class $\mathcal F_{\theta}:=\mathcal F(\tdepth,\ffdepth,\ffwidth,\clen,\dhd,m,R,\kappa)$,
    where $\tdepth$ bounds the Transformer depth,
    $\ffdepth$ and $\ffwidth$ specify the feedforward depth
    and maximum hidden width, $\clen$ is the sequence length,
    $\dhd$ is the token dimension, $m$ bounds the number of
    attention heads per shared or individual parallel block,
    and $R$ and $\kappa$ bound the output norm and parameter
    magnitudes, respectively. Details about attention, feedforward, and block constructions can be found in \ref{Definitions}.}. The proposed CAP-T represents $f$, $g_\alpha$, and $g_{\mathrm{risk}}$ using a common attention architecture with separate parameters. Each network draws on the joint characteristic matrix, allowing the relationships among firms that inform total returns, mispricing, and risk compensation to differ. The component predictions are then combined into the final return forecast. The following algorithm summarizes the estimation procedure.
	\subsection{The Estimation Algorithm}
	\label{sec:estimation_algorithm}
	
	At a forecast origin $T$, the rolling training sample consists of $\{(X_{t-1},\bm y_t):t=1,\ldots,T\}$. All model fitting and target construction use the information available through $T$, and the
	fitted models then use $X_T$ to forecast returns at $T+1$. For notational brevity, we suppress the rolling sample index throughout. Here, we set the factor count to $K$ and the local window length to $m_{\mathrm{PCA}}=Th$, where $h\in(0,1)$ and $m_{\mathrm{PCA}}\geq2$ is an integer. 
    
	\medskip
	\noindent\textbf{S1. Fit the conditional mean and construct residuals.}
	Estimate the initial transformer by least squares:
	\begin{equation*}
		\widehat f
		=\underset{F\in\mathcal F}{\arg\min}
		\sum_{t=1}^{T}\|\bm y_t-F(X_{t-1})\|_2^2.
	\end{equation*}
	For $t=1,\ldots,T$, we can compute $\widehat{\bm y}_t:=\widehat f(X_{t-1})$ and $\widehat{\bm\nu}_t:=\bm y_t-\widehat{\bm y}_t$.
	
	\medskip
	\noindent\textbf{S2. Estimate the local loading spaces.}
	For each $t=m_{\mathrm{PCA}}+1,\ldots,T$, define $\mathcal W_t:=\{t-m_{\mathrm{PCA}},\ldots,t-1\}$. Conceptually, local estimation uses the approximation $\bm\beta(X_{s-1})\approx\bm\beta(X_{t-1})$ for $s\in\mathcal W_t$, as formalized in Assumption~\ref{assumption4}. The use of temporal localization follows the local-PCA approach of \citet{SuWang2017}, applied here to residuals from the
first-stage return regression. Here, we use the quartic kernel
	\begin{equation}
		\label{eq:local_pca_kernel}
		\mathcal K(x)=
		\begin{cases}
			\displaystyle\frac{15}{16}(1-x^2)^2,&|x|\leq1,\\[3pt]
			0,&|x|>1,
		\end{cases}
		\quad
		w_{s,t}:=
		\frac{\mathcal K((s-t)/m_{\mathrm{PCA}})}
		{\sum_{j\in\mathcal W_t}\mathcal K((j-t)/m_{\mathrm{PCA}})}.
	\end{equation}
	   Note that the weights sum to one and vanish at $s=t-m_{\mathrm{PCA}}$. Then, we form the residual second moment matrix
	\begin{equation*}
		\widehat{\bm Z}_t
		:=\sum_{s\in\mathcal W_t}
		w_{s,t}\widehat{\bm\nu}_s\widehat{\bm\nu}_s^{\top}.
	\end{equation*}
	Let $\widehat{\bm v}_{1,t},\ldots,\widehat{\bm v}_{K,t}$ be orthonormal eigenvectors associated with the $K$ largest eigenvalues of $\widehat{\bm Z}_t$. Define
	\begin{equation}
		\label{eq:local_pca_loadings}
		\widehat{\bm\beta}(X_{t-1})
		:=\sqrt N[\widehat{\bm v}_{1,t},\ldots,\widehat{\bm v}_{K,t}],
		\quad
		\frac{1}{N}\widehat{\bm\beta}(X_{t-1})^{\top}
		\widehat{\bm\beta}(X_{t-1})=\bm I_K.
	\end{equation}
	Here $\widehat{\bm\beta}(X_{t-1})$ denotes the local loading estimate at date $t$. Since the window contains $m_{\mathrm{PCA}}-1$ observations with positive weights, we have $\operatorname{rank}(\widehat{\bm Z}_t)\leq \min\{N,m_{\mathrm{PCA}}-1\}$. If factor innovation estimates are required, we can compute
	\begin{equation*}
		\widehat{\bm\zeta}_{s\mid t}
		:=\frac{1}{N}\widehat{\bm\beta}(X_{t-1})^{\top}
		\widehat{\bm\nu}_s,
		\quad s\in\mathcal W_t,
	\end{equation*}
	where the index $t$ identifies the local PCA coordinates.
    
	\medskip
	\noindent\textbf{S3. Construct the component targets.}
	For each eligible date $t=m_{\mathrm{PCA}}+1,\ldots,T$, we can estimate the factor risk premia by regressing the fitted conditional mean $\widehat f(X_{t-1})$ on the estimated loadings using least squares across firms. The normalization in \eqref{eq:local_pca_loadings} yields
	\begin{equation}
		\label{eq:14}
		\widehat{\bm\lambda}(X_{t-1})
		=\frac{1}{N}\widehat{\bm\beta}(X_{t-1})^{\top}
		\widehat f(X_{t-1}).
	\end{equation}
	The risk compensation target is
	$\widetilde{g}_{\mathrm{risk}}(X_{t-1}):=\widehat{\bm\beta}(X_{t-1})\widehat{\bm\lambda}(X_{t-1})$,
	and the conditional mispricing target is
	\begin{equation*}
		\widetilde{g}_\alpha(X_{t-1})
		:=\widehat f(X_{t-1})
		-\widetilde{g}_{\mathrm{risk}}(X_{t-1}).
	\end{equation*}
	Thus, the targets satisfy
	\begin{equation}
		\label{eq:target_identities}
			\widehat f(X_{t-1})
			=\widetilde{g}_\alpha(X_{t-1})
			+\widetilde{g}_{\mathrm{risk}}(X_{t-1}),\quad \widehat{\bm\beta}(X_{t-1})^{\top}
			\widetilde{g}_\alpha(X_{t-1})=\bm 0.
	\end{equation}
	Both targets have dimension $N$ and are invariant to changes of basis within the selected loading space, whereas the estimated risk premia have dimension $K$ and depend on the coordinates.
	
	\medskip
	\noindent\textbf{S4. Learn the component pricing functions.}
	Finally, we fit separate transformers using only dates with complete residual windows. We denote the learned mispricing function by $\widehat{g}_\alpha$, so as to distinguish it from the projection estimates $\widetilde{g}_\alpha$ used as targets:
	\begin{align*}
		\widehat{g}_\alpha
		&=\underset{G\in\mathcal F}{\arg\min}
		\sum_{t=m_{\mathrm{PCA}}+1}^{T}
		\|\widetilde{g}_\alpha(X_{t-1})-G(X_{t-1})\|_2^2,\\
		\widehat{g}_{\mathrm{risk}}
		&=\underset{R\in\mathcal F}{\arg\min}
		\sum_{t=m_{\mathrm{PCA}}+1}^{T}
		\|\widetilde{g}_{\mathrm{risk}}(X_{t-1})
		-R(X_{t-1})\|_2^2.
	\end{align*}
	Note that each network is trained on its component targets using only $X_{t-1}$ as input.
	
	\medskip
	\noindent\textbf{S5. Forecast with the rolling sample.}
	For the new market state $X_T$, we can compute
	\begin{equation*}
		\widehat{\bm y}_{T+1\mid T}
		:=\widehat{g}_\alpha(X_T)+\widehat{g}_{\mathrm{risk}}(X_T).
	\end{equation*}
	This combined forecast and the direct forecast $\widehat f(X_T)$ target the same conditional mean through different estimation procedures. On the other hand, the component forecasts also provide separate investment signals for mispricing and risk compensation.
	
	\section{Theory} \label{sec:theory}
	Our theoretical analysis first exploits the intrinsic geometry of the cross-sectional state space to establish a vector-valued Transformer approximation result. The approximation
	theory is then combined with generalization bounds for dependent samples.
	Finally, we derive recovery bounds for the structural components by tracing how estimation error from the first stage propagates
	through local factor estimation and component learning. We start by stating some technical conditions\footnote{A detailed notation table can be found in \ref{Notations} of the Supplement.}.
	
	\begin{assumption}[Compact Low-Dimensional Bounded-Rank Manifold]
		\label{assumption1:manifold_p}
		For $X_t\in\mathbb{R}^{N\times P}$, assume that there exists
		$r_{\max}\in\mathbb{Z}_{+}$, with
		$r_{\max}\ll\min\{N,P\}$, such that $\sup_t\operatorname{rank}(X_t)\le r_{\max}$.
		That is, the support of $\{X_t\}$ is contained in a compact $C^2$ embedded
		submanifold
		\[
		\mathcal M
		\subseteq
		\mathcal V_{\le r_{\max}}
		:=
		\left\{
		X\in\mathbb{R}^{N\times P}:
		\operatorname{rank}(X)\le r_{\max}
		\right\}.
		\]
		Let $d:=\dim(\mathcal M)
		\le
		d_{\max}
		:=
		r_{\max}(N+P-r_{\max})$. 
		Furthermore, $\mathcal M$ has positive $\operatorname{reach}(\mathcal M)\ge\tau>0$, 
		where $\operatorname{reach}(\mathcal M)
			:=
			\sup\{
			\rho>0: \text{every }Z\in\mathbb{R}^{N\times P}
			\text{ with }\operatorname{dist}(Z,\mathcal M)<\rho,\text{has a unique nearest point in }\mathcal M
			\}$. Consequently, $\mathcal M$ admits a finite collection of uniformly
		bi-lipschitz local coordinate charts at a common geometric scale
		controlled by $\tau$.
		
	\end{assumption}
	
	Assumption~\ref{assumption1:manifold_p} states that, although each
	state $X_t$ is represented by an $N\times P$ matrix with $NP$
	entries, these entries do not vary independently. The bounded-rank
	condition restricts the state matrices to a structured subset whose
	dimension is at most
	$r_{\max}(N+P-r_{\max})$. Intuitively, a manifold is a possibly curved
	set that can be described locally by only a small number of
	coordinates. Thus, a state matrix may have many observed entries while
	its effective variation is driven by only $d$ underlying directions.
	The number $d$ is the intrinsic dimension of the state space, whereas
	$NP$ is its ambient dimension. Meanwhile, compactness ensures that the entire state space can be covered by finitely many local neighborhoods. The positive-reach condition rules
	out sharp singularities, near self-intersections, and excessively
	irregular curvature at the scale relevant for the analysis. As a
	result, each local neighborhood of $\mathcal M$ can be represented
	stably by $d$ coordinates, with distances in the local coordinates
	remaining comparable to distances between the original matrices.
	This allows the approximation problem for high-dimensional matrix
	inputs to be reduced to a finite collection of $d$-dimensional local
	problems. Consequently, the approximation rate depends primarily on
	the intrinsic dimension $d$, rather than directly on the ambient
	dimension $NP$. Similar neural-network approximation results based on
	intrinsic dimension are developed by
	\citet{manifold,Chen2019NonparametricRO}. The exact-support condition could, in principle, be relaxed
to allow states in a small neighborhood of $\mathcal M$
under suitable additional regularity conditions, following the approximate-manifold approach of
\citet{JiaoShenLinHuang2023}. In this section, we retain the current exact manifold assumption mainly for theoretical simplicity.
	
	\begin{assumption}[H\"older Continuous and Boundedness of the Pricing Components]
		\label{assumption2:target_function}
		The functions $g_\alpha,g_{\mathrm{risk}}:\mathcal M\to\mathbb R^N$
		satisfy, for some $\beta\in(0,1]$ and $H_f,R>0$,
		\[
		\|g_\alpha(X)-g_\alpha(Y)\|_2
		+
		\|g_{\mathrm{risk}}(X)-g_{\mathrm{risk}}(Y)\|_2
		\le H_f\|X-Y\|_F^\beta,
		\quad X,Y\in\mathcal M,
		\]
		and
		$
		\sup_{X\in\mathcal M}
		\left\{
		\|g_\alpha(X)\|_2+\|g_{\mathrm{risk}}(X)\|_2
		\right\}
		\le R.
		$ Since $f=g_\alpha+g_{\mathrm{risk}}$, the triangle inequality gives
		\[
		\|f(X)-f(Y)\|_2\le H_f\|X-Y\|_F^\beta,
		\quad
		\sup_{X\in\mathcal M}\|f(X)\|_2\le R.
		\]
	\end{assumption}

	Thus, the two pricing components and their sum share the
	regularity bounds used in the approximation analysis. Bi-Lipschitz local coordinates preserve their H\"older
	regularity on each chart.
	
	\begin{assumption}[State Dependence and Return Innovations]
		\label{assumption3:temporal_dynamics}
        We assume the following conditions hold.
		\leavevmode
		\begin{itemize}
			\item[(i)] \textbf{Weakly dependent states.}
			Let $Q_t$ denote the marginal distribution of $X_{t-1}$ and define
			$\bar Q_T:=T^{-1}\sum_{t=1}^TQ_t$.
			The process is uniformly
			geometrically $\beta$-mixing ($\beta_{\mathrm{mix}}$): there exist constants
			$C,c>0$, independent of $T$, such that
			\[
			\beta_{\mathrm{mix}}(k)
			:=
			\sup_{0\le t\le T-1-k}
			\beta_{\mathrm{mix}}\!\left(
			\sigma(X_0,\ldots,X_t),
			\sigma(X_{t+k},\ldots,X_{T-1})
			\right)
			\le
			C e^{-c k},
			\quad k\ge1.
			\]
			
			\item[(ii)] \textbf{Return innovations.}
			Let
			$\mathscr J_{t-1}
			:=\sigma(X_0,\ldots,X_{t-1},\bm y_1,\ldots,\bm y_{t-1})$.
			Write $\Sigma_{\zeta_t}:=\mathbb E(\bm \zeta_t\bm \zeta_t^\top)$ and
			$\Sigma_u:=\mathbb E(\bm u_t\bm u_t^\top)$; note that cross-sectional dependence within $\bm u_t$ is allowed.
			For all $s\in\mathbb R$, $\bm b\in\mathbb R^K$,
			$\bm a\in\mathbb R^N$, and $i$, $t$, we have 
			\[
			\mathbb E e^{s\bm b^\top\bm \zeta_t}
			\le e^{C_0s^2\|\bm b\|_2^2/2},
			\quad
			\mathbb E e^{s u_{it}}
			\le e^{C_0s^2/2},
			\quad
			\mathbb E\!\left[
			e^{s\bm a^\top\bm\nu_t}\mid\mathscr J_{t-1}
			\right]
			\le e^{\sigma_\nu^2s^2\|\bm a\|_2^2/2},
			\]
			where $C_0>0$ is uniform over indices and sample sizes,
			whereas $\sigma_\nu>0$ is uniform over time
			and may depend on $N$.
			
		\end{itemize}
	\end{assumption}
	
	\begin{assumption}[Slow Variation of Loadings]
		\label{assumption4}
		Let $h$ denote the temporal bandwidth used in local PCA. Across sample sizes, the effective window length $Th$ is held fixed.
		There exist $\vartheta\in(0,1]$ and a small constant
		$\delta_\beta\in(0,1)$, independent of $N$ and $T$,
		such that, almost surely,
		\[
		\frac{
			\|\bm\beta(X_{s-1})-\bm\beta(X_{t-1})\|_F
		}{\sqrt N}
		\le
		\delta_\beta
		\left(\frac{|s-t|}{Th}\right)^\vartheta,
		\quad
		1\le s,t\le T,\quad
		0<|s-t|\le Th.
		\]
	\end{assumption}
	
	\begin{assumption}[Identification, Factor Strength,
		and Idiosyncratic Covariance]
		\label{assumption5:moments_factor_strength}
		Let $K$ be fixed. There exist constants $0<c<C<\infty$
		such that the following conditions hold.
		
		\begin{itemize}
			\item[(i)] \textbf{Identification.}
			Almost surely, for every $t$, we have $\bm\beta(X_{t-1})^\top g_\alpha(X_{t-1})=\bm0$.

			\item[(ii)] \textbf{Uniform bounds.}
			The loading rows and risk premia satisfy
			\[
			\sup_{X\in\mathcal M}
			\max_{1\le i\le N}
			\|\bm\beta_{i\cdot}(X)\|_2\le C,
			\quad
			\sup_t\|\bm\lambda(X_{t-1})\|_2\le C.
			\]
			The bound $R$ in Assumption~\ref{assumption2:target_function} satisfies $R^2\le CN$.
			
			\item[(iii)] \textbf{Factor strength.}
			Almost surely, uniformly over $t$,
			\[
			cI_K
			\preceq
			\frac1N\bm\beta(X_{t-1})^\top\bm\beta(X_{t-1})
			\preceq
			CI_K,
			\quad
			cI_K\preceq\Sigma_v\preceq CI_K.
			\]
			
			\item[(iv)] \textbf{Weak idiosyncratic covariance.}
			The idiosyncratic covariance satisfies
			$\|\Sigma_u\|_{\mathrm{op}}\le C$.
		\end{itemize}
	\end{assumption}
	
	Assumptions~\ref{assumption3:temporal_dynamics}--%
	\ref{assumption5:moments_factor_strength} provide the
	innovation, slow-variation, identification, and factor
	strength conditions used by the residual based local PCA.
	The loading matrix is identified up to a nonsingular
	transformation of factor coordinates.
	The idiosyncratic covariance may exhibit cross-sectional
	dependence, but its operator norm is uniformly bounded,
	whereas the nonzero eigenvalues of the factor covariance
	component are of order $N$.
	
	\subsection{Transformer Function Approximation and Generalization Theory}
	In this section, we give the approximation and generalization theory for the proposed method.
	\begin{theorem} [$N$-Output Manifold Approximation Theorem] \label{thm:1}
		Let $\tau, R, H_f > 0$, $0 < \beta \le 1$, $d, N,P \in \mathbb{N}$ and $\mathcal{M}$ satisfies Assumption \ref{assumption1:manifold_p}. For any $\epsilon \in (0, 1)$, there exists a transformer neural network class $\mathcal{F}(L_T, L_{\mathrm{FFN}}, w_{\mathrm{FFN}}, l, d_{\mathrm{embd}}, m, R, \kappa)$ with parameters
		\begin{equation*}
			\begin{aligned}
				&L_T = O(\log(d)), & &L_{\mathrm{FFN}} = O(\log(\epsilon^{-1})), & &w_{\mathrm{FFN}} = O(1), & &l = O(d\epsilon^{-\frac{d}{\beta}}) \\
				& d_{\mathrm{embd}} = O(1), & &m = O(d\epsilon^{-\frac{d}{\beta}}), & & \kappa = O(d^2\epsilon^{-\frac{2d}{\beta}}), & & L_{\mathrm{para}} < L_T,
			\end{aligned}
		\end{equation*}
		\textit{where $O(\cdot)$ hides terms in $C_{\mathcal{M}}, N, P, H_{f},R, \tau$, such that for $f$ satisfying Assumption \ref{assumption2:target_function}, we have}
		\begin{equation*}
			\|\mathrm F_\theta-f\|_{L^\infty(\mathcal M;\mathbb R^N)}
			:=
			\sup_{X\in\mathcal M}
			\|\mathrm F_\theta(X)-f(X)\|_2
			\le \epsilon,
		\end{equation*}
		for some $\mathrm{F}_\theta \in \mathcal{F}(L_T, L_{\mathrm{FFN}}, w_{\mathrm{FFN}}, l, d_{\mathrm{embd}}, m, R, \kappa)$. The same class also contains separate approximations to
		$g_\alpha$ and $g_{\mathrm{risk}}$, each with uniform error
		at most $\epsilon$. 
	\end{theorem}
	Theorem~\ref{thm:1} establishes the approximation foundation of the
	analysis. Although the input state $X$ is an $N\times P$ matrix, its
	support is restricted to the $d$-dimensional manifold $\mathcal M$.
	The Transformer first represents the high-dimensional input through
	local coordinates on $\mathcal M$ and then uses these low-dimensional
	representations to approximate the $N$-dimensional conditional return
	function. Consequently, the dependence of the approximation complexity
	on the target accuracy $\epsilon$ is governed by the intrinsic
	dimension $d$, while the ambient dimensions $N$ and $P$ enter only
	through the multiplicative constants. Based on the approximation result, we establish the following Theorem~\ref{thm:2} by deriving the corresponding statistical error rate.

	\begin{theorem}[Statistical Error under Temporal Dependence]
		\label{thm:2}
		Suppose that Assumptions~\ref{assumption1:manifold_p},
		\ref{assumption2:target_function}, and
		\ref{assumption3:temporal_dynamics}(i)--(ii) hold.
		Write $\bm y_t=f(X_{t-1})+\bm\nu_t$, where
		$f=g_\alpha+g_{\mathrm{risk}}$ is the conditional mean
		return function.
		Let $\varphi^2$ denote a deterministic statistical error term satisfying
		$\varphi^2=\widetilde O(
		NPd^2\epsilon^{-d/\beta}/T)$,
		\[
		\|\widehat f-f\|_{L^2(\bar Q_T;\mathbb R^N)}^2
		+
		\frac1T\sum_{t=1}^T
		\|\widehat f(X_{t-1})-f(X_{t-1})\|_2^2
		=O_p(\epsilon^2+\varphi^2).
		\]
		In particular, choosing
		$\epsilon=T^{-\beta/(2\beta+d)}$ gives $\|\widehat f-f\|_{L^2(\bar Q_T;\mathbb R^N)}^2
		=
		\widetilde O_p(
		NPd^2T^{-2\beta/(2\beta+d)}
		)$, where $\widetilde O_p$ suppresses logarithmic factors.
	\end{theorem}
	
	Theorem~\ref{thm:2} builds on the approximation result in
    Theorem~\ref{thm:1}, with the required regularity of $f$
    following from Assumption~\ref{assumption2:target_function}.
    It bounds the initial regression error under temporal
    dependence and time-varying state distributions. It controls
    both the $L^2(\bar Q_T;\mathbb R^N)$ error\footnote{
    For $G:\mathcal M\to\mathbb R^N$,
    $\|G\|_{L^2(\bar Q_T;\mathbb R^N)}^2
    :=\int_{\mathcal M}\|G(X)\|_2^2\,d\bar Q_T(X)$,
    where $\bar Q_T=T^{-1}\sum_{t=1}^TQ_t$
    and $Q_t$ is the marginal distribution of $X_{t-1}$.}
    and the average squared error over observed states in
    probability, providing the basis for subsequent factor
    and component estimation. For fixed dimensions and
    geometric quantities, the convergence exponent depends
    on the smoothness $\beta$ and intrinsic dimension $d$,
    while the bound also reflects network complexity and
    the variability of return innovations.

	For the in-sample analysis, we can extend the local PCA
	construction to all dates $t=1,\ldots,T$.
	For $0<h\le1/2$, $Th\in\mathbb N$, and $Th\ge2$, define
	\[
	\mathcal W_t
	:=
	\begin{cases}
		\{1,\ldots, Th\},
		&1\le t\le Th,\\
		\{t- Th,\ldots,t-1\},
		& Th<t\le T,
	\end{cases}
	\]
	and the initial windows use nearby observations within
	the training sample.
	All windows contain $Th$ observations
	and satisfy $|s-t|\le Th$ for $s\in\mathcal W_t$.
	For this theoretical construction, the S3 targets
	are formed at all $T$ dates, and the S4 regressions
	use all these targets. Note that at each out-of-sample forecast origin, the entire
	construction uses only the available training sample.
	
	\begin{theorem}[Local Loading Recovery]
		\label{thm:3}
		Suppose that the conditions of Theorem~\ref{thm:2}
		and Assumptions~\ref{assumption4}--
		\ref{assumption5:moments_factor_strength} hold.
		Let $1\le K<N$.
		Construct $\widehat{\bm\beta}(X_{t-1})$
		by residual-based local PCA using the windows
		defined above and the quartic kernel in
		\eqref{eq:local_pca_kernel}.
		Write $\eta^2:=N/(Th)+N\delta_\beta^2+N^{-1}$.
		There exist nonsingular matrices
		$H_t$ satisfying
		$N^{-1}H_t^\top
		\bm\beta(X_{t-1})^\top
		\bm\beta(X_{t-1})H_t=I_K$
		and
		$\|H_t\|_{\mathrm{op}}+
		\|H_t^{-1}\|_{\mathrm{op}}\le C$,
		such that
		\[
		\frac1T\sum_{t=1}^T
		\left\|
		\widehat{\bm\beta}(X_{t-1})
		-\bm\beta(X_{t-1})H_t
		\right\|_F^2
		=O_p\left(
		\epsilon^2+\varphi^2+\eta^2
		\right).
		\]
	\end{theorem}
	
	Theorem~\ref{thm:3} controls the average loading error in probability
	up to a change of factor coordinates.
	The bound accounts for first-stage estimation,
	local second-moment fluctuations, temporal variation
	of the loadings, and idiosyncratic covariance.
	The boundary windows satisfy the same error bound
	as the historical windows.
	Under Assumption~\ref{assumption4}, local variation
	contributes $N\delta_\beta^2$, which need not vanish
	when $\delta_\beta$ is fixed.
	
	Then, the S4 regressions use the class $\mathcal F$ specified
	in Theorem~\ref{thm:2}. And by Theorem~\ref{thm:1}, this class contains separate
	approximations to $g_\alpha$ and $g_{\mathrm{risk}}$,
	each with uniform error at most $\epsilon$.
	
	\begin{theorem}[Statistical Error Bounds for Conditional Pricing Components]
		\label{thm:4}
		Suppose that the conditions of
		Theorem~\ref{thm:3} hold, then we have
	\begin{equation*}
	\frac1T\sum_{t=1}^T
	\|\widehat g_\alpha(X_{t-1})-g_\alpha(X_{t-1})\|_2^2
	+
	\|\widehat g_\alpha-g_\alpha\|_{L^2(\bar Q_T;\mathbb R^N)}^2
	=O_p(\epsilon^2+\varphi^2+\eta^2)
    \end{equation*}
     \begin{equation*}
    \frac1T\sum_{t=1}^T
	\|\widehat g_{\mathrm{risk}}(X_{t-1})
	-g_{\mathrm{risk}}(X_{t-1})\|_2^2
	+
	\|\widehat g_{\mathrm{risk}}-g_{\mathrm{risk}}\|_
	{L^2(\bar Q_T;\mathbb R^N)}^2
	=O_p(\epsilon^2+\varphi^2+\eta^2).
	\end{equation*}
		Consequently, their sum satisfies
		\[
		\|\widehat{g}_\alpha
		+\widehat{g}_{\mathrm{risk}}-f\|_
		{L^2(\bar Q_T;\mathbb R^N)}^2
		=O_p(\epsilon^2+\varphi^2+\eta^2).
		\]
	\end{theorem}
	Theorem~\ref{thm:4} traces how the initial estimation and
    loading space recovery affect the estimated pricing
    components and their sum. The bounds account for
    approximation, statistical estimation, and local PCA
    errors through $\epsilon^2$, $\varphi^2$, and $\eta^2$,
    respectively, both over the observed states and under
    the average marginal distribution $\bar Q_T$.
    The following corollary relates estimation of the
    combined pricing function to return prediction.
	\begin{corollary}[Mean Squared Forecast Error Decomposition]
		\label{cor:prediction_error}
		Under the conditions of Theorem~\ref{thm:4},
		let $\widehat g_\alpha^{(t)}$ and
		$\widehat g_{\mathrm{risk}}^{(t)}$ be fitted using
		information available at forecast origin $t$, and define
		$ \widehat{\bm y}_{t+1\mid t}
		:= \widehat g_\alpha^{(t)}(X_t) + \widehat g_{\mathrm{risk}}^{(t)}(X_t).
		$
		Then we have
		\[ 
		\mathbb E
		\|\bm y_{t+1}-\widehat{\bm y}_{t+1\mid t}\|_2^2 =
		\mathbb E\|\bm\nu_{t+1}\|_2^2 +
		\mathbb E \|\widehat{\bm y}_{t+1\mid t}-f(X_t)\|_2^2.
		\]
		The identity also holds after averaging over the
		evaluation dates.
	\end{corollary}
    The corollary separates forecast error into the conditional mean estimation error and irreducible innovation variance.
    The former arises from estimating the sum of the two
    pricing components, whose recovery is studied in
    Theorem~\ref{thm:4}; while the latter reflects factor innovations
    and idiosyncratic shocks with zero conditional mean given
    the information available at the forecast origin.
    This variation remains unpredictable under that
    information set, even when the pricing functions are
    known.
	
	\section{Simulation and Empirical Analysis}
	\label{sec:evidence}
	
	In this section, we assess CAP-T through simulation and application to
	U.S. equities. The simulation examines recovery of the
	pricing components and loading space when the underlying
	structure is known. The empirical analysis evaluates return
	prediction, the information captured by the estimated
	components, and their use in portfolio allocation.
	
	In both analyses, we compare CAP-T with IPCA \citep{IPCA}
	and DNN-varying \citep{Fan1}. IPCA specifies factor loadings
	as linear functions of firm characteristics. DNN-varying
	combines neural networks estimated separately at each date
	with local PCA and subsequent component regressions.
	Its pricing functions use each firm's own characteristics,
	whereas CAP-T incorporates information across firms. We consider two CAP-T variants with ReLU and Softmax attention, respectively.
	ReLU attention follows our theoretical specification,
	while Softmax provides a conventional alternative.
	Softmax assigns nonnegative weights that sum to one, whereas ReLU sets nonpositive scores to zero without normalizing
	the remaining weights. These coefficients aggregate
	predictive information, making normalization an architectural
	choice. We hold the remaining architectural and training settings
	fixed when comparing the two attention mechanisms.
	For each CAP-T variant, the networks for total expected
	returns, conditional mispricing, and risk compensation
	share an architectural template but are estimated
	separately. Both analyses consider one to three layers
	for CAP-T and DNN-varying.

    \subsection{Simulation}
	\label{simulation}
	
	We examine the recovery of conditional mispricing,
	risk compensation, and the factor loading space under
	smoothly varying exposures. Each simulated panel contains
	$N=120$ firms observed over $T=600$ periods, with $K=5$
	factors. The state matrix varies over time through a calendar index and
	includes 18 characteristics and an intercept, giving $P=20$:
    \begin{equation*}
    \begin{aligned}
        (X_{t-1})_{i1}
        &=1,\qquad (X_{t-1})_{i,11}=t/T,\\
        (X_{t-1})_{i,j+1+\mathbf{1}\{j\geq10\}}
        &=\sqrt{2}\cos\!\left[\frac{\pi j(i-1/2)}{N}\right],
        \qquad j=1,\ldots,18.
    \end{aligned}
    \end{equation*}
    The first loading is constant,
    $[\bm\beta(X_{t-1})]_{i1}=1$,
    while the remaining loadings rotate smoothly:
    \begin{equation*}
        [\bm\beta(X_{t-1})]_{ij}
        =(X_{t-1})_{ij}\cos(1.6jt/T)
        +(X_{t-1})_{i,j+4}\sin(1.6jt/T),
        \qquad j=2,\ldots,5.
    \end{equation*}
    $g_\alpha$ and $\lambda$ vary with the calendar index:
    \begin{equation*}
        \begin{aligned}
            [g_\alpha(X_{t-1})]_i
            &=0.005\{1+0.1\sin(2\pi t/T)\}(X_{t-1})_{i,10},\\
            \bm\lambda(X_{t-1})
            &=\left\{1+0.05\sin\!\left[\frac{\pi}{2}(X_{t-1})_{1,11}\right]\right\}
            (0.005,\,0.003,\,-0.002,\,0.002,\,0.001)^\top.
        \end{aligned}
    \end{equation*}
    The construction satisfies
    $\bm\beta(X_{t-1})^\top\bm\beta(X_{t-1})/N=\bm I_5$ and
    $\bm\beta(X_{t-1})^\top g_\alpha(X_{t-1})=\bm0$.
    Factor innovations $\bm \zeta_t$ are independent across dates and components, with
    zero-mean Gaussian distributions and standard deviations $(0.020,
    0.018, 0.016, 0.014, 0.012)$. The errors $u_{it}$ are iid $N(0,0.001^2)$
    and independent of $\bm\zeta_t$.
    We generate ten panels with independent innovation paths and common
    pricing functions, estimating all models at $K=5$.
    
    The first 80\% of each panel is reserved for initial estimation and the
    remaining 20\% for out-of-sample evaluation. CAP-T and IPCA use
    360-period rolling training windows, six-period parameter updates, and
    one-period-ahead forecasts. CAP-T constructs component targets from
    local PCA of residuals at the preceding 12 dates. A quartic kernel with
    bandwidth nine gives positive weight to the eight most recent residuals.
    The comparison includes Softmax and ReLU CAP-T and DNN-varying, each
    with one to three layers, and IPCA. For an evaluation period $\mathcal T$, estimation accuracy is measured by
    $
    \operatorname{RMSE}(\widehat{\bm c},\bm c;\mathcal T)
    =\{
    \sum_{t\in\mathcal T}
    \|\widehat{\bm c}_t-\bm c_t\|_2^2/N|\mathcal T|\}^{1/2}.
    $
    Predictions are evaluated against observed $\bm y$, while estimated
    components and decomposition targets are compared with their true values.
    We also report $R^2$ for recombined predictions, using fitted values
    in sample and forecasts out of sample. Following \citet{YuWangSamworth2015}, let
    $\theta_{j,t}=\arccos\{\sigma_j(U_t^\top\widehat U_t)\}$
    be the principal angles between the true and estimated loading spaces,
    where $U_t$ and $\widehat U_t$ are orthonormal bases and $\sigma_j$
    denotes the $j$th largest singular value.
    We use a normalized version of the projection distance in
    \citet{EdelmanAriasSmith1998}, averaged over the evaluation period:
    \begin{equation*}
    	\overline d_{\mathrm{proj}}(\mathcal T)
    	=\frac{1}{|\mathcal T|}\sum_{t\in\mathcal T}
    	\left\{\frac{1}{K}\sum_{j=1}^{K}\sin^2\theta_{j,t}\right\}^{1/2}
    	=\frac{1}{|\mathcal T|}\sum_{t\in\mathcal T}
    	\frac{\|U_tU_t^\top-\widehat U_t\widehat U_t^\top\|_F}{\sqrt{2K}}.
    \end{equation*}
    This basis-invariant distance ranges from zero for identical spaces
    to one for orthogonal spaces. All statistics are computed within each simulated panel and then
    averaged across ten panels. For DNN-varying, label and loading diagnostics
    use contemporaneous observations, whereas forecasts use previously
    estimated networks. For IPCA, label RMSE evaluates the mispricing
    estimate obtained by alternating least squares.

    Table~\ref{tab:sim-k5} shows that all six CAP-T specifications outperform
    DNN-varying and IPCA on the five error measures and Total $R^2$ in
    both samples. Outside the estimation sample, one-layer Softmax achieves
    the lowest return RMSE and the highest Total $R^2$,
    at 4.8601\%, while two-layer ReLU gives the lowest mispricing RMSE.
    Three-layer Softmax gives the lowest loading-space error,
    0.1169, while two-layer Softmax gives the lowest risk-compensation RMSE,
    $2.5911\times10^{-3}$. Thus, additional depth does not uniformly improve
    performance. For every CAP-T specification, learning the mispricing
    function reduces RMSE relative to the constructed targets.

    \begin{table}[!htbp]
	\centering
	\caption{Recovery and prediction with five factors}
	\label{tab:sim-k5}
    	\begingroup
    	\small\setstretch{1.08}
    	\setlength{\tabcolsep}{3pt}\renewcommand{\arraystretch}{1.08}
    	\begin{tabular*}{\textwidth}{@{\extracolsep{\fill}}lc*{6}{r}@{}}
    		\toprule
    		& & \multicolumn{4}{c}{RMSE ($\times 10^3$)}
    		& & \multicolumn{1}{c}{$R^2$ (\%)} \\
    		\cmidrule(lr){3-6}\cmidrule(lr){8-8}
    		Model & Layers
    		& \multicolumn{1}{c}{Returns}
    		& \multicolumn{1}{c}{Mispricing}
    		& \multicolumn{1}{c}{Risk}
    		& \multicolumn{1}{c}{Labels}
    		& \multicolumn{1}{c}{$\overline d_{\mathrm{proj}}$}
    		& \multicolumn{1}{c}{Total} \\
    		\midrule
    		\multicolumn{8}{@{}l}{
    			\textit{In sample}} \\
    		\addlinespace[2pt]
    		\multirow{3}{*}{\shortstack[l]{CAP-T\\(ReLU)}}
    		& 1 & 35.9126 & 1.0044 & 1.9866 & \textbf{1.6872} & \textbf{0.0973} & 5.8683 \\
    		& 2 & 35.8735 & 0.9328 & 2.4123 & 1.7793 & 0.1019 & 6.0729 \\
    		& 3 & \textbf{35.8493} & \textbf{0.8918} & 2.7917 & 1.7758 & 0.1014 & \textbf{6.1993} \\
    		\addlinespace[5pt]
    		\multirow{3}{*}{\shortstack[l]{CAP-T\\(Softmax)}}
    		& 1 & 35.9200 & 0.9995 & \textbf{1.9133} & 1.7012 & 0.0983 & 5.8291 \\
    		& 2 & 35.8952 & 0.9499 & 2.0860 & 1.7511 & 0.1001 & 5.9593 \\
    		& 3 & 35.8797 & 0.9298 & 2.3178 & 1.7688 & 0.1014 & 6.0400 \\
    		\addlinespace[5pt]
    		\multirow{3}{*}{DNN-varying}
    		& 1 & 36.5745 & 5.1210 & 4.6059 & 4.7242 & 0.7655 & 2.3648 \\
    		& 2 & 36.5582 & 5.0652 & 4.5957 & 4.8593 & 0.7737 & 2.4516 \\
    		& 3 & 36.5494 & 5.0216 & 4.5972 & 4.6831 & 0.7738 & 2.4994 \\
    		\addlinespace[5pt]
    		IPCA & \textemdash & 40.9091 & 8.3983 & 22.0522 & 8.3983 & 0.5290 & $<0$ \\
    		\midrule
    		\multicolumn{8}{@{}l}{
    			\textit{Out of sample}} \\
    		\addlinespace[2pt]
    		\multirow{3}{*}{\shortstack[l]{CAP-T\\(ReLU)}}
    		& 1 & 36.5379 & 1.1902 & 2.7354 & 2.1551 & 0.1241 & 4.8527 \\
    		& 2 & 36.5474 & \textbf{1.0551} & 2.7373 & 2.1423 & 0.1211 & 4.8050 \\
    		& 3 & 36.5804 & 1.2557 & 3.1518 & 2.3012 & 0.1278 & 4.6300 \\
    		\addlinespace[5pt]
    		\multirow{3}{*}{\shortstack[l]{CAP-T\\(Softmax)}}
    		& 1 & \textbf{36.5364} & 1.1294 & 2.6121 & 2.3036 & 0.1293 & \textbf{4.8601} \\
    		& 2 & 36.5460 & 1.2904 & \textbf{2.5911} & 2.1392 & 0.1228 & 4.8105 \\
    		& 3 & 36.5497 & 1.2424 & 2.6656 & \textbf{2.0223} & \textbf{0.1169} & 4.7917 \\
    		\addlinespace[5pt]
    		\multirow{3}{*}{DNN-varying}
    		& 1 & 37.3789 & 5.4006 & 6.0211 & 5.5403 & 0.7869 & 0.4200 \\
    		& 2 & 37.2842 & 5.1155 & 5.6546 & 5.6774 & 0.7999 & 0.9227 \\
    		& 3 & 37.3029 & 5.0864 & 5.8033 & 5.2973 & 0.7933 & 0.8244 \\
    		\addlinespace[5pt]
    		IPCA & \textemdash & 37.6078 & 2.4039 & 9.6094 & 2.4039 & 0.6055 & $<0$ \\
    		\bottomrule
    	\end{tabular*}
    	\endgroup
    \end{table}

\begin{figure}[!htbp]
    	\centering
    	\includegraphics[width=\textwidth]{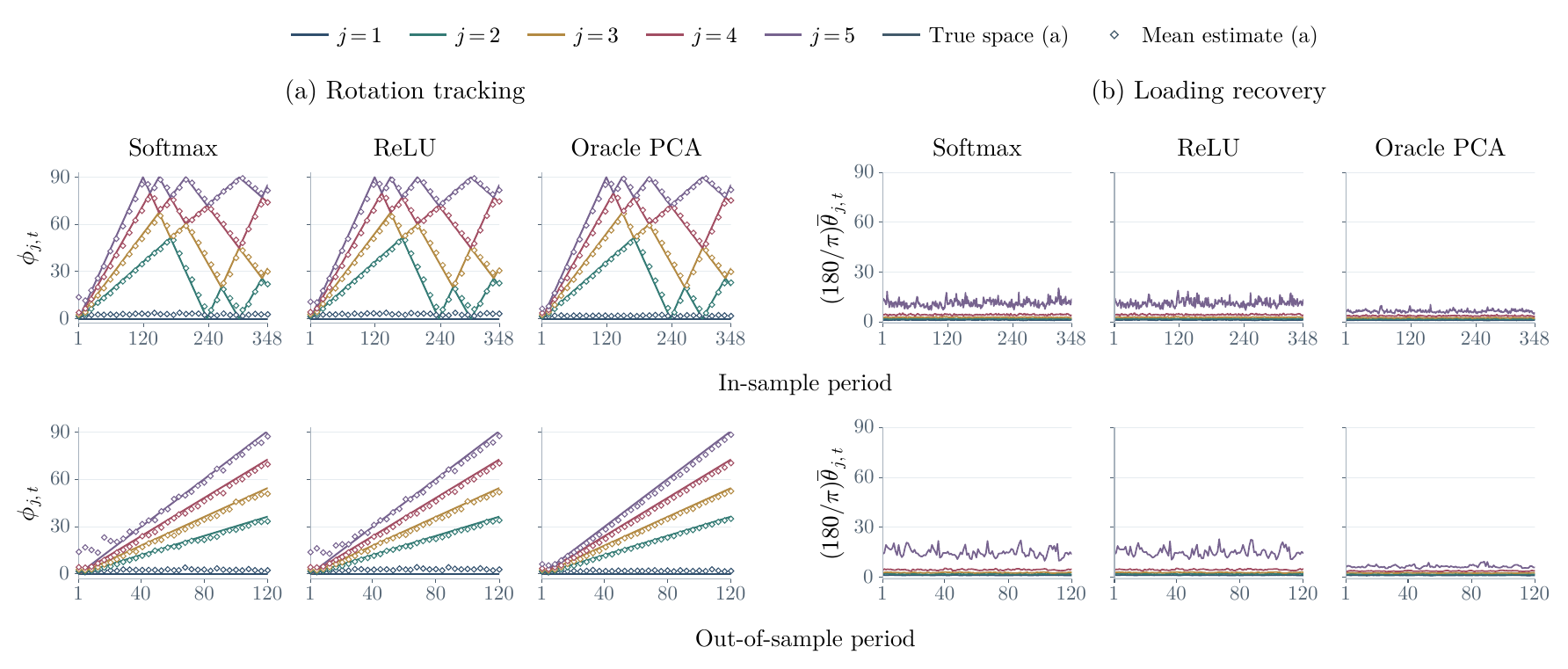}
    	\caption{Tracking and recovery of the loading space.
    	In the left three columns, lines show
    	$\phi_{j,t}=(180/\pi)\arccos\{\sigma_j(U_*^\top U_t)\}$
    	and hollow diamonds show mean estimated counterparts, where
    	$U_*$ is the true basis at the first date of each row.
    	The right three columns show $(180/\pi)\overline\theta_{j,t}$;
    	smaller values indicate better recovery.
    	The upper and lower rows show the estimation and out-of-sample
    	periods.}
    	\label{fig:sim-loading-comparison}
    \end{figure}

    Figure~\ref{fig:sim-loading-comparison} shows that both CAP-T variants
    track the rotation of the loading space in sample and out of sample.
    In the left three columns, the estimated angles closely follow the
    true trajectories. In the right three columns, the first four ordered
    principal angles remain small for both variants, while the fifth is
    higher. Softmax has a lower average fifth angle than ReLU out of sample,
    indicating slightly better alignment with the true loading space.
    These patterns support accurate tracking and recovery beyond the
    initial estimation sample.

	\subsection{Empirical Analysis}
	\label{real data}
	We apply the models introduced in Section~\ref{simulation} to a monthly panel spanning 1965-2020. Returns come from CRSP, and accounting information is from Compustat, following the data sources used by \citet{Fan1,data1,IPCA}. The inputs comprise 150 firm characteristics drawn primarily from \citet{JensenKellyPedersen2023}. The underlying universe consists of U.S. common stocks listed on the NYSE, AMEX, and NASDAQ. To accommodate reporting lags, accounting characteristics for fiscal years ending in calendar year $t-1$ are matched to returns from July of year $t$ through June of year $t+1$.
	
	After aligning firms across sources and applying the sample filters, we retain a balanced panel of 167 firms with 670 monthly observations and no missing returns. Characteristics observed from February 1965 through November 2020 are paired with excess returns in the following month, from March 1965 through December 2020. The first 536 observations form the initial estimation sample. The remaining 134 months, from November 2009 through December 2020, form the common evaluation period.

	Using the same attention mechanisms and network depths as in the simulation, we assess return prediction, component prediction, and portfolio performance in the observed data. For a forecast $C_{t\mid t-1}$, we report the uncentered coefficient of determination
	\begin{equation*}
		R^2(C)
		=1-\frac{\sum_{t\in\mathcal T_{\mathrm{OOS}}}
			\|\bm y_t-C_{t\mid t-1}\|_2^2}
		{\sum_{t\in\mathcal T_{\mathrm{OOS}}}\|\bm y_t\|_2^2},
	\end{equation*}
	where $\mathcal T_{\mathrm{OOS}}$ contains the 134 evaluation months. We evaluate the Total forecast $\widehat{\bm y}_{t\mid t-1}$ and its two components, $\widehat{g}_\alpha(X_{t-1})$ and $\widehat{g}_{\mathrm{risk}}(X_{t-1})$. Risk compensation excludes realized factor innovations. Each component is scored against full realized excess returns, so the component scores are not additive. Scores are computed for each fitted member before averaging. CAP-T uses three independently fitted pipelines; DNN-varying uses three forecast members sharing the same decomposition targets; and IPCA provides a deterministic forecast. All $R^2$ values are reported in percent, with negative values displayed as $<0$.
	
	Table~\ref{tab:rd_schedules} examines monthly, quarterly, and semiannual parameter updating. The interval $M$ controls the re-estimation frequency, while the forecast horizon remains one month. Reported ranges cover all factor counts and network depths within each model family. CAP-T maintains positive scores for the Total forecast and both components throughout the grid. Its advantage over the benchmarks persists across updating schedules, factor counts, and network depths. Here we only report semiannual updating, i.e., $M=6$ to save space, while \ref{app:rd_updating_intervals} reports the other schedules.
	
	\begin{table}[!htbp]
		\centering
		\caption{Predictive performance across updating schedules}
		\label{tab:rd_schedules}
		\begingroup
		\small\setstretch{1.08}
		\setlength{\tabcolsep}{3pt}\renewcommand{\arraystretch}{1.12}
		\begin{tabular*}{\textwidth}{@{\extracolsep{\fill}}lc*{7}{r}@{}}
			\toprule
			& & & \multicolumn{3}{c}{Minimum $R^2$ (\%)} & \multicolumn{3}{c}{Maximum $R^2$ (\%)} \\
			\cmidrule(lr){4-6}\cmidrule(lr){7-9}
			Model & Layers & $M$ & Total & Risk & Mispricing & Total & Risk & Mispricing \\
			\midrule
			\multirow{3}{*}{\shortstack[l]{CAP-T\\(ReLU)}} & \multirow{3}{*}{1--3} & 1 & 1.5215 & 0.9622 & 0.7784 & 1.6309 & 1.2820 & 1.1289 \\
			&  & 3 & 1.5438 & 0.9651 & 0.8034 & 1.6611 & 1.2839 & 1.1618 \\
			&  & 6 & 1.5680 & 0.9689 & 0.8176 & 1.6815 & 1.2964 & 1.1866 \\
			\addlinespace[5pt]
			\multirow{3}{*}{\shortstack[l]{CAP-T\\(Softmax)}} & \multirow{3}{*}{1--3} & 1 & 1.5756 & 0.9685 & 0.7899 & 1.6710 & 1.3017 & 1.1399 \\
			&  & 3 & 1.6001 & 1.0093 & 0.8256 & 1.6942 & 1.3123 & 1.1791 \\
			&  & 6 & \textbf{1.6140} & \textbf{1.0169} & \textbf{0.8435} & \textbf{1.7161} & \textbf{1.3201} & \textbf{1.2017} \\
			\addlinespace[5pt]
			\multirow{3}{*}{DNN-varying} & \multirow{3}{*}{1--3} & 1 & $<0$ & 0.0740 & $<0$ & $<0$ & 0.1944 & $<0$ \\
			&  & 3 & 0.3106 & 0.4994 & $<0$ & 0.4842 & 0.6278 & $<0$ \\
			&  & 6 & 0.4984 & 0.6893 & $<0$ & 0.6246 & 0.7881 & 0.0413 \\
			\addlinespace[5pt]
			\multirow{3}{*}{IPCA} & \multirow{3}{*}{\textemdash} & 1 & $<0$ & 0.0294 & $<0$ & $<0$ & 0.1791 & $<0$ \\
			&  & 3 & $<0$ & 0.0448 & $<0$ & $<0$ & 0.1814 & $<0$ \\
			&  & 6 & $<0$ & 0.0543 & $<0$ & $<0$ & 0.1899 & $<0$ \\
			\bottomrule
		\end{tabular*}
		\par\vspace{5pt}
		\begin{minipage}{\textwidth}
			\small\setstretch{1.1}
			\textit{Notes:} $M$ is the parameter updating interval in months; all forecasts are for the following month. Ranges span $K=1,\ldots,7$ and, for the neural networks, one to three layers, after averaging scores across fitted members. All models use the same 134 evaluation months. Risk and Mispricing are each evaluated against full realized excess returns. Bold identifies the highest endpoint in each numeric column.
		\end{minipage}
		\endgroup
	\end{table}
	
	Then, Table~\ref{tab:rd_attention} separates the roles of attention, depth, and factor dimension. CAP-T retains its predictive advantage across configurations. Softmax generally provides the strongest Total forecasts, while ReLU remains effective throughout the grid. Factor dimension has a clearer effect on component prediction than on the Total forecast. Risk compensation scores rise with $K$ for every CAP-T specification, while mispricing scores decline but remain positive. This pattern is consistent with the changing definition of mispricing as the factor space expands. The Total forecast remains comparatively stable. For example, as $K$ increases from one to seven, the two layer Softmax model's risk compensation $R^2$ rises from 1.0541\% to 1.3201\%, while its Total forecast $R^2$ changes only modestly, from 1.7161\% to 1.6448\%.
	The attention mechanisms exhibit complementary strengths: Softmax with two layers leads risk compensation prediction, while ReLU with three layers performs particularly well for mispricing at larger factor counts. At $K=7$, the three layer ReLU model achieves the highest mispricing $R^2$ of 0.8989\%, while the two layer Softmax model leads risk compensation prediction at 1.3201\%. These results motivate allowing the component networks to learn different relationships among firms.
	
	\begin{table}[!htbp]
		\centering
		\caption{Predictive accuracy by attention mechanism, depth, and factor dimension}
		\label{tab:rd_attention}
		\begingroup
		\small\setstretch{1.05}
		\setlength{\tabcolsep}{3pt}\renewcommand{\arraystretch}{1.12}
		\begin{tabular*}{\textwidth}{@{\extracolsep{\fill}}lc*{7}{r}@{}}
			\toprule
			& & \multicolumn{7}{c}{Number of factors, $K$} \\
			\cmidrule(lr){3-9}
			Model & Layers & \multicolumn{1}{c}{1} & \multicolumn{1}{c}{2} & \multicolumn{1}{c}{3} & \multicolumn{1}{c}{4} & \multicolumn{1}{c}{5} & \multicolumn{1}{c}{6} & \multicolumn{1}{c}{7} \\
			\midrule
			\multicolumn{9}{@{}l}{\textit{Panel A. Total return forecast}} \\
			\addlinespace[2pt]
			\multirow{3}{*}{\shortstack[l]{CAP-T\\(ReLU)}} & 1 & 1.6815 & 1.6240 & 1.6142 & 1.6316 & 1.6284 & 1.6334 & 1.6400 \\
			& 2 & 1.6034 & 1.5933 & 1.5821 & 1.5680 & 1.5693 & 1.5775 & 1.5916 \\
			& 3 & 1.6434 & 1.6576 & 1.6411 & 1.6408 & 1.6289 & 1.6235 & 1.6215 \\
			\addlinespace[5pt]
			\multirow{3}{*}{\shortstack[l]{CAP-T\\(Softmax)}} & 1 & 1.6785 & 1.6865 & 1.6421 & 1.6345 & 1.6297 & 1.6202 & 1.6140 \\
			& 2 & \textbf{1.7161} & \textbf{1.6931} & 1.6702 & \textbf{1.6712} & \textbf{1.6652} & 1.6531 & 1.6448 \\
			& 3 & 1.6960 & 1.6860 & \textbf{1.6706} & 1.6662 & 1.6615 & \textbf{1.6532} & \textbf{1.6454} \\
			\addlinespace[5pt]
			\multirow{3}{*}{DNN-varying} & 1 & 0.4984 & 0.5595 & 0.5650 & 0.5673 & 0.5676 & 0.5700 & 0.5751 \\
			& 2 & 0.5280 & 0.5349 & 0.5358 & 0.5386 & 0.5425 & 0.5469 & 0.5504 \\
			& 3 & 0.6096 & 0.6151 & 0.6246 & 0.6145 & 0.6238 & 0.6238 & 0.6218 \\
			\addlinespace[5pt]
			IPCA & \textemdash & $<0$ & $<0$ & $<0$ & $<0$ & $<0$ & $<0$ & $<0$ \\
			\midrule
			\multicolumn{9}{@{}l}{\textit{Panel B. Risk compensation}} \\
			\addlinespace[2pt]
			\multirow{3}{*}{\shortstack[l]{CAP-T\\(ReLU)}} & 1 & 1.0229 & 1.0937 & 1.1397 & 1.1955 & 1.2234 & 1.2597 & 1.2964 \\
			& 2 & 0.9689 & 1.0915 & 1.1275 & 1.1538 & 1.1859 & 1.2220 & 1.2668 \\
			& 3 & 0.9982 & 1.1392 & 1.1700 & 1.2049 & 1.2250 & 1.2458 & 1.2764 \\
			\addlinespace[5pt]
			\multirow{3}{*}{\shortstack[l]{CAP-T\\(Softmax)}} & 1 & 1.0185 & 1.1656 & 1.1807 & 1.2142 & 1.2434 & 1.2658 & 1.2928 \\
			& 2 & \textbf{1.0541} & \textbf{1.1719} & \textbf{1.2099} & \textbf{1.2526} & \textbf{1.2805} & \textbf{1.2979} & \textbf{1.3201} \\
			& 3 & 1.0169 & 1.1424 & 1.1858 & 1.2237 & 1.2526 & 1.2731 & 1.2953 \\
			\addlinespace[5pt]
			\multirow{3}{*}{DNN-varying} & 1 & 0.7881 & 0.7842 & 0.7548 & 0.7540 & 0.7487 & 0.7580 & 0.7538 \\
			& 2 & 0.7541 & 0.7228 & 0.6913 & 0.6906 & 0.6898 & 0.6893 & 0.6936 \\
			& 3 & 0.7112 & 0.7451 & 0.7368 & 0.7181 & 0.7366 & 0.7290 & 0.7208 \\
			\addlinespace[5pt]
			IPCA & \textemdash & 0.1899 & 0.1104 & 0.1532 & 0.1105 & 0.0888 & 0.0543 & 0.1163 \\
			\midrule
			\multicolumn{9}{@{}l}{\textit{Panel C. Conditional mispricing}} \\
			\addlinespace[2pt]
			\multirow{3}{*}{\shortstack[l]{CAP-T\\(ReLU)}} & 1 & 1.1801 & 1.0742 & 1.0234 & 0.9805 & 0.9438 & 0.9081 & 0.8696 \\
			& 2 & 1.1223 & 1.0136 & 0.9655 & 0.9258 & 0.8903 & 0.8568 & 0.8176 \\
			& 3 & 1.1866 & 1.0845 & \textbf{1.0388} & \textbf{1.0045} & \textbf{0.9714} & \textbf{0.9388} & \textbf{0.8989} \\
			\addlinespace[5pt]
			\multirow{3}{*}{\shortstack[l]{CAP-T\\(Softmax)}} & 1 & 1.1942 & \textbf{1.0881} & 1.0373 & 0.9943 & 0.9557 & 0.9192 & 0.8772 \\
			& 2 & \textbf{1.2017} & 1.0875 & 1.0303 & 0.9839 & 0.9441 & 0.9071 & 0.8651 \\
			& 3 & 1.1655 & 1.0524 & 0.9984 & 0.9544 & 0.9166 & 0.8820 & 0.8435 \\
			\addlinespace[5pt]
			\multirow{3}{*}{DNN-varying} & 1 & $<0$ & $<0$ & $<0$ & $<0$ & $<0$ & $<0$ & $<0$ \\
			& 2 & $<0$ & $<0$ & $<0$ & $<0$ & $<0$ & $<0$ & $<0$ \\
			& 3 & 0.0413 & $<0$ & $<0$ & $<0$ & $<0$ & $<0$ & $<0$ \\
			\addlinespace[5pt]
			IPCA & \textemdash & $<0$ & $<0$ & $<0$ & $<0$ & $<0$ & $<0$ & $<0$ \\
			\bottomrule
		\end{tabular*}
		\par\vspace{5pt}
		\begin{minipage}{\textwidth}
			\small\setstretch{1.1}
			\textit{Notes:} Out-of-sample $R^2$ in percent under semiannual parameter updating ($M=6$). Scores for the neural network models are averaged across three random seeds; IPCA is deterministic. The total forecast is the sum of predicted mispricing and risk compensation. Corresponding results for $M=1$ and $M=3$ are reported in \ref{app:rd_updating_intervals}.
		\end{minipage}
		\endgroup
	\end{table}

    \subsubsection{Economic Performance Evaluation}

    We evaluate the investment value of CAP-T through portfolio allocation
    based on predicted total returns and stock selection based on conditional
    alpha. Both strategies are fully invested and long-only. The long-only restriction is motivated by its role in mitigating estimation risk and avoiding short-selling frictions
    \citep{JagannathanMa2003,DAvolio2002,DuffieGarleanuPedersen2002}. Allocation rules are identical
    across models and factor counts within each strategy. Portfolios rebalance
    monthly, and performance is evaluated over the 134 months from November 2009 to December 2020.
    And reported Sharpe ratios are obtained by averaging the annualized ratios computed from net monthly excess returns.\footnote{
    The net annualized Sharpe ratio is
    $\sqrt{12}\,\bar r_{\mathrm{net}}/s(r_{\mathrm{net}})$, where
    $r_{t+1,\mathrm{net}}
    =\bm w_t^\top\bm y_{t+1}-0.002\,\mathrm{TO}_t$
    and
    $\mathrm{TO}_t=\frac12\|\bm w_t-\bm w_{t-1}\|_1$.
    Here, $\bm w_t$ denotes target portfolio weights, initial turnover is
    $0.5$, and $\bar r_{\mathrm{net}}$ and $s(r_{\mathrm{net}})$ are the
    sample mean and sample standard deviation of monthly net excess returns.}

    The first strategy uses predicted total returns as expected returns
    in the mean--variance framework of \citet{Markowitz1952}, maximizing the forecast return-to-volatility ratio, with the portfolio fully invested and each asset weight no more than 12\%.
    Performance is compared with equal-weighted (EW) and value-weighted
    (VW) portfolios of all 167 firms, with VW weights based on
    previous-month market capitalization. The same turnover-based
    transaction costs apply to all portfolios.

    Table~\ref{tab:rd_forecast_sr} shows that CAP-T achieves higher net
    Sharpe ratios than all DNN-varying and IPCA specifications at every
    factor count. Across architectures and factor dimensions, CAP-T's
    net Sharpe ratios range from 1.0756 to 1.1402, compared with 0.8276
    and 0.7341 for the full-universe EW and VW benchmarks.
    The highest value is attained with six factors.
    Performance varies modestly across factor counts within each
    architecture, indicating that CAP-T's allocation advantage extends
    across the specifications considered.

    \begin{table}[!htbp]
        \centering
        \caption{Forecast-based mean--variance portfolios: net Sharpe ratios}
        \label{tab:rd_forecast_sr}
        \begingroup
        \small\setstretch{1.08}
        \setlength{\tabcolsep}{3pt}
        \renewcommand{\arraystretch}{1.12}
        \begin{tabular*}{\textwidth}{@{\extracolsep{\fill}}lc*{7}{r}@{}}
            \toprule
            & & \multicolumn{7}{c}{Number of factors, $K$} \\
            \cmidrule(lr){3-9}
            Model & Layers & 1 & 2 & 3 & 4 & 5 & 6 & 7 \\
            \midrule
            \multirow{3}{*}{\shortstack[l]{CAP-T\\(ReLU)}}
            & 1 & \textbf{1.1382} & 1.1245 & 1.1101 & 1.1094 & 1.1122 & 1.1089 & 1.1091 \\
            & 2 & 1.0756 & 1.0829 & 1.0759 & 1.0865 & 1.0817 & 1.0847 & 1.0807 \\
            & 3 & 1.1117 & 1.1284 & \textbf{1.1322} & \textbf{1.1372} & \textbf{1.1364} & \textbf{1.1402} & \textbf{1.1361} \\
            \addlinespace[5pt]
            \multirow{3}{*}{\shortstack[l]{CAP-T\\(Softmax)}}
            & 1 & 1.1146 & 1.1241 & 1.1128 & 1.1139 & 1.1094 & 1.1087 & 1.0978 \\
            & 2 & 1.0861 & 1.1014 & 1.0981 & 1.1039 & 1.1070 & 1.1026 & 1.0930 \\
            & 3 & 1.1227 & \textbf{1.1307} & 1.1206 & 1.1172 & 1.1147 & 1.1086 & 1.0963 \\
            \addlinespace[5pt]
            \multirow{3}{*}{DNN-varying}
            & 1 & 1.0388 & 1.0419 & 1.0375 & 1.0389 & 1.0434 & 1.0316 & 1.0328 \\
            & 2 & 1.0037 & 1.0010 & 1.0052 & 1.0135 & 1.0193 & 0.9985 & 0.9973 \\
            & 3 & 1.0686 & 1.0577 & 1.0448 & 1.0599 & 1.0641 & 1.0644 & 1.0692 \\
            \addlinespace[5pt]
            IPCA & \textemdash & 0.7909 & 0.7513 & 0.7391 & 0.7704 & 0.8097 & 0.8635 & 0.8794 \\
            \bottomrule
        \end{tabular*}
        \par\vspace{5pt}
        \begin{minipage}{\textwidth}
            \small\setstretch{1.1}
            \textit{Notes:} The full-universe EW and VW benchmarks have
            net Sharpe ratios of 0.8276 and 0.7341, respectively.
        \end{minipage}
        \endgroup
    \end{table}

    The second strategy examines whether conditional alpha identifies
    attractive stocks within the long-only mandate. Adapting the
    forecast-based decile-sorting approach of \citet{GuKellyXiu2020},
    we rank firms by predicted conditional alpha and retain only the
    highest decile. This gives 17 of the 167 firms, with the decile size
    rounded up. Portfolio weights are proportional to predicted alpha
    within this group; all selected alpha forecasts are positive in the
    evaluated sample. The strategy therefore uses the ranking to select
    stocks and signal magnitudes to allocate capital, without a short leg.

    Table~\ref{tab:rd_alpha_sr} demonstrates the stock-selection value
    of CAP-T's conditional alpha. For $K=2,\ldots,7$, every CAP-T
    architecture delivers a higher net Sharpe ratio than all
    DNN-varying and IPCA specifications, while two architectures
    also lead at $K=1$. The strongest portfolio achieves a net
    Sharpe ratio of 0.8710 with seven factors. These results support the investment value of conditional alpha
    across factor dimensions and network architectures.

    \begin{table}[!htbp]
        \centering
        \caption{Top-decile conditional-alpha portfolios: net Sharpe ratios}
        \label{tab:rd_alpha_sr}
        \begingroup
        \small\setstretch{1.08}
        \setlength{\tabcolsep}{3pt}
        \renewcommand{\arraystretch}{1.12}
        \begin{tabular*}{\textwidth}{@{\extracolsep{\fill}}lc*{7}{r}@{}}
            \toprule
            & & \multicolumn{7}{c}{Number of factors, $K$} \\
            \cmidrule(lr){3-9}
            Model & Layers & 1 & 2 & 3 & 4 & 5 & 6 & 7 \\
            \midrule
            \multirow{3}{*}{\shortstack[l]{CAP-T\\(ReLU)}}
            & 1 & 0.7457 & 0.7890 & \textbf{0.8300} & 0.7951 & \textbf{0.8538} & 0.8268 & 0.8041 \\
            & 2 & 0.7122 & 0.7528 & 0.8289 & 0.8316 & 0.8361 & 0.8534 & 0.8362 \\
            & 3 & \textbf{0.7925} & \textbf{0.8211} & 0.8022 & 0.8339 & 0.8483 & \textbf{0.8576} & \textbf{0.8710} \\
            \addlinespace[5pt]
            \multirow{3}{*}{\shortstack[l]{CAP-T\\(Softmax)}}
            & 1 & 0.7721 & 0.8169 & 0.8279 & \textbf{0.8551} & 0.8518 & 0.8375 & 0.8184 \\
            & 2 & 0.7570 & 0.7469 & 0.7478 & 0.7381 & 0.7244 & 0.7425 & 0.7272 \\
            & 3 & 0.7692 & 0.7266 & 0.7149 & 0.7645 & 0.7644 & 0.7555 & 0.7584 \\
            \addlinespace[5pt]
            \multirow{3}{*}{DNN-varying}
            & 1 & 0.7703 & 0.7223 & 0.6898 & 0.6927 & 0.6869 & 0.6679 & 0.6576 \\
            & 2 & 0.5263 & 0.5217 & 0.5654 & 0.5453 & 0.5596 & 0.5411 & 0.5630 \\
            & 3 & 0.5301 & 0.5610 & 0.5902 & 0.5923 & 0.5681 & 0.5950 & 0.5529 \\
            \addlinespace[5pt]
            IPCA & \textemdash & 0.6780 & 0.7098 & 0.6935 & 0.7161 & 0.6700 & 0.6462 & 0.6395 \\
            \bottomrule
        \end{tabular*}
        \endgroup
    \end{table}

    Table~\ref{tab:rd_average_finance} reports financial statistics for
    $K=5$. Panel A considers forecast-based allocation and Panel B
    conditional-alpha selection. 
    \begin{table}[!htbp]
        \centering
        \caption{Financial performance with five factors}
        \label{tab:rd_average_finance}
        \begingroup
        \small\setstretch{1.08}
        \setlength{\tabcolsep}{3pt}
        \renewcommand{\arraystretch}{1.12}
        \begin{tabular*}{\textwidth}{@{\extracolsep{\fill}}lc*{7}{r}@{}}
            \toprule
            Model & Layers & SR & AR & Vol. & MDD & TO & Cost & CR \\
            & & & (\%) & (\%) & (\%) & & (\%) & (\%) \\
            \midrule
            \multicolumn{9}{@{}l}{\textit{Panel A. Forecast-based mean--variance portfolios}} \\
            \addlinespace[2pt]
            \multirow{3}{*}{\shortstack[l]{CAP-T\\(ReLU)}}
            & 1 & 1.112 & 12.344 & 11.099 & 18.338 & 3.314 & 0.663 & 268.254 \\
            & 2 & 1.082 & 12.188 & 11.264 & 18.804 & 3.036 & 0.607 & 261.628 \\
            & 3 & \textbf{1.136} & 12.391 & \textbf{10.904} & 17.154 & \textbf{2.812} & \textbf{0.562} & 271.111 \\
            \addlinespace[5pt]
            \multirow{3}{*}{\shortstack[l]{CAP-T\\(Softmax)}}
            & 1 & 1.109 & 12.598 & 11.356 & 18.377 & 4.204 & 0.841 & 277.672 \\
            & 2 & 1.107 & 12.517 & 11.306 & 19.061 & 3.639 & 0.728 & 274.740 \\
            & 3 & 1.115 & 12.588 & 11.297 & 17.844 & 3.474 & 0.695 & \textbf{277.697} \\
            \addlinespace[5pt]
            \multirow{3}{*}{DNN-varying}
            & 1 & 1.043 & 11.407 & 10.935 & 18.054 & 5.094 & 1.019 & 232.641 \\
            & 2 & 1.019 & 11.137 & 10.925 & 16.847 & 4.862 & 0.972 & 224.269 \\
            & 3 & 1.064 & 11.809 & 11.095 & \textbf{16.262} & 4.245 & 0.849 & 247.534 \\
            \addlinespace[5pt]
            IPCA & \textemdash & 0.810 & \textbf{12.978} & 16.027 & 29.146 & 9.808 & 1.962 & 267.356 \\
            \midrule
            \multicolumn{9}{@{}l}{\textit{Panel B. Top-decile conditional-alpha portfolios}} \\
            \addlinespace[2pt]
            \multirow{3}{*}{\shortstack[l]{CAP-T\\(ReLU)}}
            & 1 & \textbf{0.854} & 12.961 & 15.070 & 26.446 & 9.460 & 1.892 & 283.232 \\
            & 2 & 0.836 & 13.036 & 15.756 & 31.035 & \textbf{9.401} & \textbf{1.880} & 272.606 \\
            & 3 & 0.848 & 11.249 & \textbf{13.330} & \textbf{21.983} & 9.602 & 1.920 & 218.971 \\
            \addlinespace[5pt]
            \multirow{3}{*}{\shortstack[l]{CAP-T\\(Softmax)}}
            & 1 & 0.852 & \textbf{14.046} & 16.569 & 30.395 & 9.986 & 1.997 & \textbf{310.322} \\
            & 2 & 0.724 & 12.077 & 16.650 & 33.967 & 9.934 & 1.987 & 235.231 \\
            & 3 & 0.764 & 12.562 & 16.627 & 31.718 & 9.633 & 1.927 & 248.246 \\
            \addlinespace[5pt]
            \multirow{3}{*}{DNN-varying}
            & 1 & 0.687 & 12.107 & 17.619 & 29.241 & 9.877 & 1.975 & 226.785 \\
            & 2 & 0.560 & 9.073 & 16.395 & 31.115 & 9.912 & 1.982 & 137.412 \\
            & 3 & 0.568 & 9.774 & 17.228 & 34.965 & 9.902 & 1.980 & 152.593 \\
            \addlinespace[5pt]
            IPCA & \textemdash & 0.670 & 13.948 & 20.819 & 47.778 & 10.107 & 2.021 & 269.103 \\
            \bottomrule
        \end{tabular*}
        \par\vspace{5pt}
        \begin{minipage}{\textwidth}
            \small\setstretch{1.1}
            \textit{Notes:} AR denotes annualized mean net excess return;
            Vol., annualized volatility; MDD, maximum drawdown;
            TO, annualized one-way turnover; Cost, annualized trading cost;
            and CR, cumulative net excess return.
            CR and MDD are computed from an index that compounds net monthly
            excess returns, starting at one. TO is expressed in portfolio
            notional units; all other measures except SR are percentages.
            Bold denotes the highest SR, AR, and CR and the lowest
            Vol., MDD, TO, and Cost within each panel.
        \end{minipage}
        \endgroup
    \end{table}

    Forecast-based allocation delivers higher net Sharpe ratios for
    CAP-T across all specifications. Its portfolios combine annualized
    net excess returns of 12.188--12.598\% with volatility of
    10.904--11.356\%, while turnover and trading costs remain below
    those of every competing model. These results highlight CAP-T's
    ability to translate return forecasts into portfolios with favorable
    risk-adjusted performance and moderate trading requirements.
    
    Conditional-alpha selection also delivers consistently higher net
    Sharpe ratios for CAP-T than for DNN-varying and IPCA, with annualized
    net excess returns reaching 14.046\% and cumulative net excess returns
    reaching 310.322\%. The two strategies serve complementary investment
    objectives: forecast-based allocation balances expected returns
    against their covariance structure, while alpha selection concentrates
    capital in firms with the strongest estimated mispricing.
    At five factors, forecast-based allocation achieves higher net
    Sharpe ratios and lower volatility, drawdown, and turnover for every
    CAP-T specification, illustrating the benefits of incorporating
    covariance information into portfolio construction.
    
    \subsubsection{Economic structure in cross-firm attention}
    \label{sec:rd_attention_structure}

    Having evaluated CAP-T's predictive and investment performance, we examine whether its attention representations capture economically meaningful relationships among firms. We select Softmax networks with strong task-specific out-of-sample $R^2$: three layers for first-stage prediction, two for risk compensation, and one for conditional alpha. We fix $K=5$, semiannual updating, covering 167 firms over all 134 out-of-sample months.
    
    We apply spectral clustering to the firm representations produced by each network's final combined multihead attention output, after projection and before residual and feedforward operations. Spectral clustering treats observations as nodes of a similarity graph and uses its eigenvectors to identify groups with strong internal connections relative to their connections with other groups \citep{vonLuxburg2007}. This approach summarizes the geometry of the learned representations without imposing industry classifications. For each month, we construct Gaussian affinities from pairwise Euclidean distances, using the median positive distance as bandwidth, and normalize the affinity matrix before averaging it over time. Following the normalized spectral clustering procedure, we compute the eigenvectors associated with the smallest eigenvalues of the normalized graph Laplacian, normalize the rows of the resulting embedding, and apply $k$-means. We select the number of groups from two to ten using the largest gap between consecutive eigenvalues, an eigengap heuristic described by \citet{vonLuxburg2007}. This criterion selects two groups for each of the three networks. We interpret these partitions through observed firm characteristics and assess selected pairwise associations against documentary evidence and characteristic similarity.

    \begin{figure}[!htb]
    \centering
    \includegraphics[width=\textwidth]{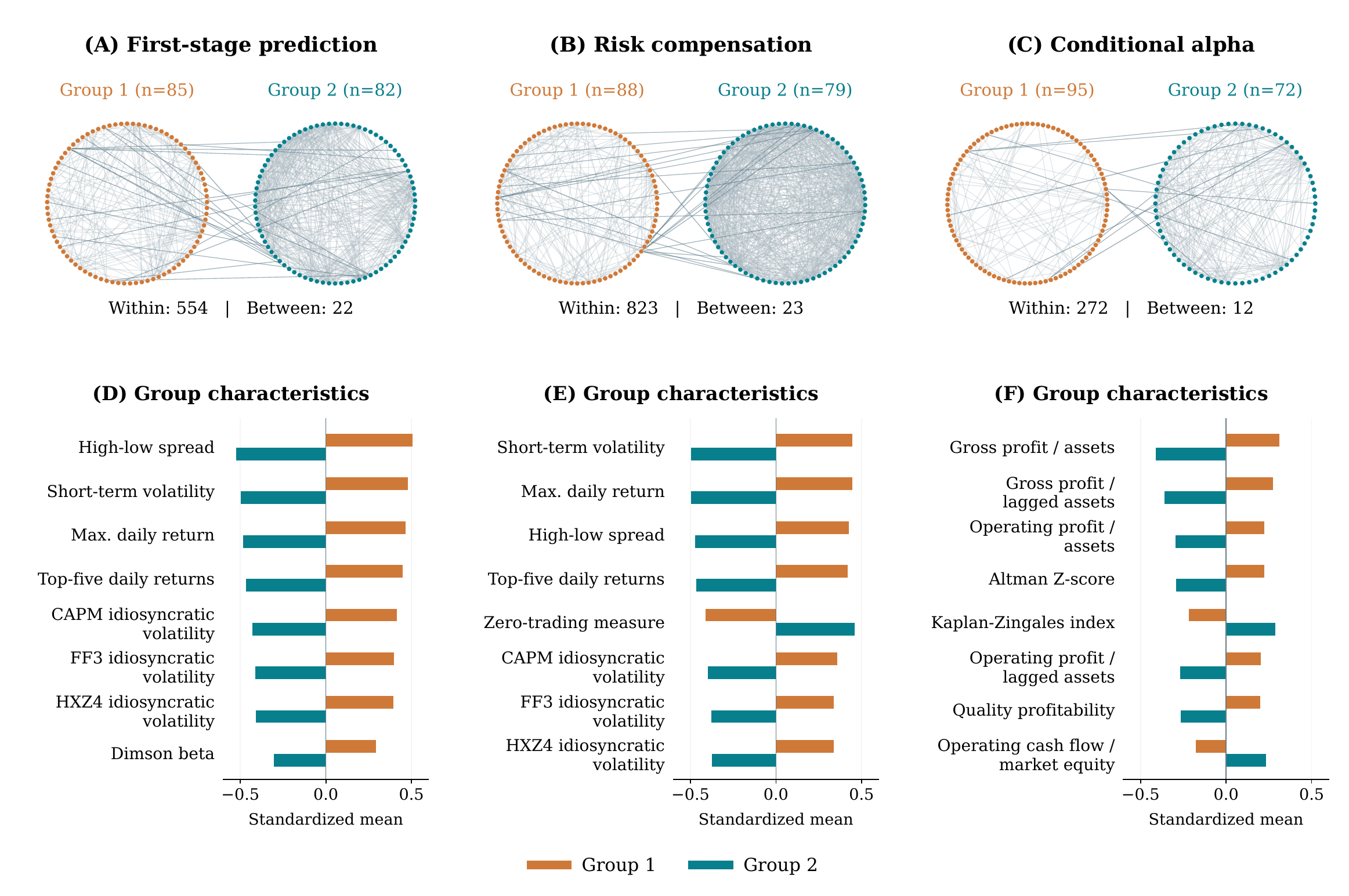}
    \caption{Cross-firm attention networks and their economic profiles}
    \label{fig:rd_attention_structure}
    \begin{minipage}{\textwidth}\setstretch{1.1}\small
    \textit{Notes:} Panels A--C display all 167 firms, colored by cluster. Edges connect pairs with co-clustering scores $\geq 0.30$; scores are the smaller of the two half-sample averages across monthly partitions with 2--10 groups. Thicker lines indicate higher scores; darker lines connect different clusters. Each node represents a firm; the circular layout is used only for visualization. Panels D--F report the eight largest absolute group-mean differences in monthly cross-sectionally standardized characteristics averaged within groups over time.
    \end{minipage}
    \end{figure}

    Figure~\ref{fig:rd_attention_structure} shows that the learned representations organize firms along economically interpretable dimensions. In the first-stage network, the groups contain 85 and 82 firms; the risk network produces a similar split of 88 and 79. Group 1 in both networks has greater short-term volatility, more extreme positive daily returns, and higher idiosyncratic volatility. The first stage partition also distinguishes the high-low spread measure. These patterns connect the information exchanged by the predictors to common risk and trading conditions. Alpha instead separates 95 and 72 firms. Its first group has higher gross and operating profitability and higher Altman $Z$-scores, whereas its second group has a higher Kaplan-Zingales index. The contrast gives the alpha representation a distinct interpretation in terms of operating strength and financing constraints.

    We measure pairwise association through co-clustering frequencies, drawing on evidence accumulation and consensus clustering \citep{FredJain2005,MontiEtAl2003}. For each pair, we average the same group indicators across monthly spectral partitions with two to ten groups, separately within each 67-month half, and take the smaller average to emphasize temporal persistence. For visualization, we retain scores of at least 0.30 among all 167 firms, using the same threshold within and between groups. This yields 554 within-group and 22 between-group links for the first stage, 823 and 23 for risk, and 272 and 12 for alpha. Cross group links capture recurring associations across months and group counts beyond the two-group partition of the time-averaged graph.

    To assess the economic content of these associations, we retain each firm's three highest scoring partners among the other 166 firms. Counting each unordered pair once yields 402, 409, and 373 distinct pairs for the first-stage, risk, and alpha networks, respectively. We examine annual reports, regulatory filings, official corporate releases, and court records for evidence of economic links, including named counterparties, shared customers, and common product or end markets. Pairs without documentary support are assessed for similarity in observed characteristics; those satisfying neither criterion are recorded as having insufficient evidence. We also assess similarity in firms' 150 characteristics, requiring it to persist across both sample halves. Table~\ref{tab:rd_attention_evidence} shows that a majority of selected pairs in each network are supported by documentary evidence or characteristic similarity.

    \begin{table}[!htbp]
    	\centering
    	\caption{Economic interpretation of firm associations}
    	\label{tab:rd_attention_evidence}
    	\begingroup
    	\small
    	\setlength{\tabcolsep}{6pt}
    	\renewcommand{\arraystretch}{1.15}
    	\begin{tabular*}{\textwidth}{@{\extracolsep{\fill}}lrrr@{}}
    		\toprule
    		& First stage & Risk & Alpha \\
    		\midrule
    		Documented links          & 128 & 124 & 102 \\
    		Characteristic similarity & 142 & 157 & 134 \\
    		Unresolved                & 132 & 128 & 137 \\
    		\midrule
    		Total pairs               & 402 & 409 & 373 \\
    		Share with supporting evidence (\%)
    		& 67.2 & 68.7 & 63.3 \\
    		\bottomrule
    	\end{tabular*}
    	\par\vspace{5pt}
    	\begin{minipage}{\textwidth}
    		\small\setstretch{1.1}
    		\textit{Notes:} 
    		Documentary support takes precedence over characteristic similarity;
    		the reported share combines these two categories.
    		Unresolved pairs meet neither criterion in our ex post review.
    	\end{minipage}
    	\endgroup
    \end{table}

    Table~\ref{tab:rd_attention_examples} illustrates five reports supported pairs per network. The first stage recovers named customers, ventures, competitors, and financial service contracts. Meanwhile, risk adds engine production and gas-processing links. For alpha, the Abbott-Pfizer pair illustrates therapeutic overlap shared with the first stage selection, while Dover-SPX adds a distinct link through heating and cooling products. These cases give the component representations distinct economic interpretations and show how they recover complementary company relationships.

    \begin{table}[!htbp]
    	\centering
    	\caption{Selected firm associations and documentary evidence}
    	\label{tab:rd_attention_examples}
    	\begingroup
    	\small
    	\setstretch{1.05}
    	\setlength{\tabcolsep}{5pt}
    	\renewcommand{\arraystretch}{1.12}
    	
    	\begin{tabularx}{\textwidth}{
    			@{}
    			>{\raggedright\arraybackslash}p{3.3cm}
    			>{\raggedright\arraybackslash}X
    			>{\centering\arraybackslash}p{1.35cm}
    			@{}
    		}
    		\toprule
    		Firm pair & Documentary evidence & Network \\
    		\midrule
    		
    		Boeing / Curtiss-Wright
    		& Boeing is a principal customer of Curtiss-Wright's
    		commercial-aircraft business.
    		\href{https://www.sec.gov/Archives/edgar/data/26324/000002632421000007/cw-20201231.htm}{[1]}
    		& F \\
    		
    		\addlinespace[3pt]
    		Boeing / Lockheed Martin
    		& Partners in the United Launch Alliance joint venture.
    		\href{https://www.sec.gov/Archives/edgar/data/12927/000001292721000011/ba-20201231.htm}{[1]}
    		& F \\
    		
    		\addlinespace[3pt]
    		HP / Xerox
    		& Xerox is a named competitor in HP's printing business.
    		\href{https://www.sec.gov/Archives/edgar/data/47217/000004721720000045/hpq-20201031.htm}{[1]}
    		& F \\
    		
    		\addlinespace[3pt]
    		Harris / Northrop Grumman
    		& Northrop Grumman is among Harris's principal competitors.
    		\href{https://www.sec.gov/Archives/edgar/data/202058/000020205819000090/hrs628201910-k.htm}{[1]}
    		& F \\
    		
    		\addlinespace[3pt]
    		Disney / Wells Fargo
    		& Wells Fargo Bank serves as an indenture trustee for Disney.
    		\href{https://www.sec.gov/Archives/edgar/data/1744489/000174448920000197/dis-20201003.htm}{[1]}
    		& F \\
    		
    		\addlinespace[6pt]
    		GE / Raytheon Technologies
    		& GE and Pratt \& Whitney participate in an engine
    		and aftermarket joint venture.
    		\href{https://www.sec.gov/Archives/edgar/data/40545/000004054521000011/ge-20201231.htm}{[1]}
    		& R \\
    		
    		\addlinespace[3pt]
    		EQT / Williams
    		& Williams Ohio Valley Midstream provides gas processing
    		and NGL marketing under contract with EQT.
    		\href{https://www.sec.gov/Archives/edgar/data/33213/000003321321000006/eqt-20201231.htm}{[1]}
    		& F, R \\
    		
    		\addlinespace[3pt]
    		Hershey / Altria
    		& McLane is a major customer of both firms.
    		\href{https://www.sec.gov/Archives/edgar/data/47111/000004711121000007/hsy-20201231.htm}{[1]},
    		\href{https://www.sec.gov/Archives/edgar/data/764180/000076418021000037/mo-20201231.htm}{[2]}
    		& F, R \\
    		
    		\addlinespace[3pt]
    		ITT / Parker-Hannifin
    		& Overlapping fluid-handling and motion-control products.
    		\href{https://www.sec.gov/Archives/edgar/data/216228/000021622821000005/itt-20201231.htm}{[1]},
    		\href{https://www.sec.gov/Archives/edgar/data/76334/000007633420000099/ph630202010-k.htm}{[2]}
    		& R \\
    		
    		\addlinespace[3pt]
    		MDU Resources / ONEOK
    		& Related natural-gas pipeline and midstream operations.
    		\href{https://www.sec.gov/Archives/edgar/data/67716/000006771621000007/mdu-20201231.htm}{[1]},
    		\href{https://www.sec.gov/Archives/edgar/data/1039684/000103968421000015/oke-20201231.htm}{[2]}
    		& R \\
    		
    		\addlinespace[6pt]
    		Abbott / Pfizer
    		& Overlapping cardiovascular and metabolic
    		pharmaceutical markets.
    		\href{https://www.sec.gov/Archives/edgar/data/1800/000110465921025751/abt-20201231x10k.htm}{[1]},
    		\href{https://www.sec.gov/Archives/edgar/data/78003/000007800321000038/pfe-20201231.htm}{[2]}
    		& F, A \\
    		
    		\addlinespace[3pt]
    		Dover / SPX
    		& Overlapping heating, cooling and refrigeration
    		equipment businesses.
    		\href{https://www.sec.gov/Archives/edgar/data/29905/000002990521000010/dov-20201231.htm}{[1]},
    		\href{https://www.sec.gov/Archives/edgar/data/88205/000008820521000009/spxc-20201231.htm}{[2]}
    		& A \\
    		
    		\addlinespace[3pt]
    		Baxter / HEICO
    		& Baxter's medical devices and HEICO's electronics
    		serve medical applications.
    		\href{https://www.sec.gov/Archives/edgar/data/10456/000162828021001867/bax-20201231.htm}{[1]},
    		\href{https://www.sec.gov/Archives/edgar/data/46619/000004661920000078/hei-20201031.htm}{[2]}
    		& A \\
    		
    		\addlinespace[3pt]
    		Spire / Sempra
    		& Spire and Sempra's SoCalGas operate regulated
    		natural-gas distribution businesses.
    		\href{https://www.sec.gov/Archives/edgar/data/1126956/000156459020054448/sr-10k_20200930.htm}{[1]},
    		\href{https://www.sec.gov/Archives/edgar/data/1032208/000103220821000007/sre-20201231.htm}{[2]}
    		& A \\
    		
    		\addlinespace[3pt]
    		Honeywell / ITT
    		& Both supply equipment and components for aircraft.
    		\href{https://www.sec.gov/Archives/edgar/data/773840/000077384021000015/hon-20201231.htm}{[1]},
    		\href{https://www.sec.gov/Archives/edgar/data/216228/000021622821000005/itt-20201231.htm}{[2]}
    		& A \\
    		
    		\bottomrule
    	\end{tabularx}
    	
    	\par\vspace{5pt}
    	\begin{minipage}{\textwidth}
    		\footnotesize\setstretch{1.1}
    		\textit{Notes:} F, R, and A denote inclusion in the first-stage,
    		risk, and alpha Top3 pair sets. Bracketed links point to supporting
    		filings within each row.
    	\end{minipage}
    	\endgroup
    \end{table}

    \subsubsection{Characteristic importance across pricing components}
    \label{sec:rd_characteristic_importance}
    
    Having examined cross-firm relationships, we turn to the characteristics driving risk, alpha, and total return predictions, where total returns are predicted by summing risk and alpha. Importance is compared across ReLU and Softmax networks with one to three layers, using five factors, six-month updates, and 134 out-of-sample months for 167 firms.
    
    Following the zero-out approach of \citet{GuKellyXiu2020}, we measure importance by the decrease in out-of-sample $R^2$ after setting a standardized characteristic to zero, holding the fitted parameters for each rolling forecast fixed. For total Y, we perturb both component inputs and sum the resulting predictions. All three predictors are evaluated against realized returns. Within each architecture and predictor, we average the signed decreases over three independently fitted members, truncate negative averages at zero, and normalize the resulting scores to sum to one across all 150 characteristics.
    
\begin{figure}[!htbp]
    \centering
    \includegraphics[width=\textwidth]{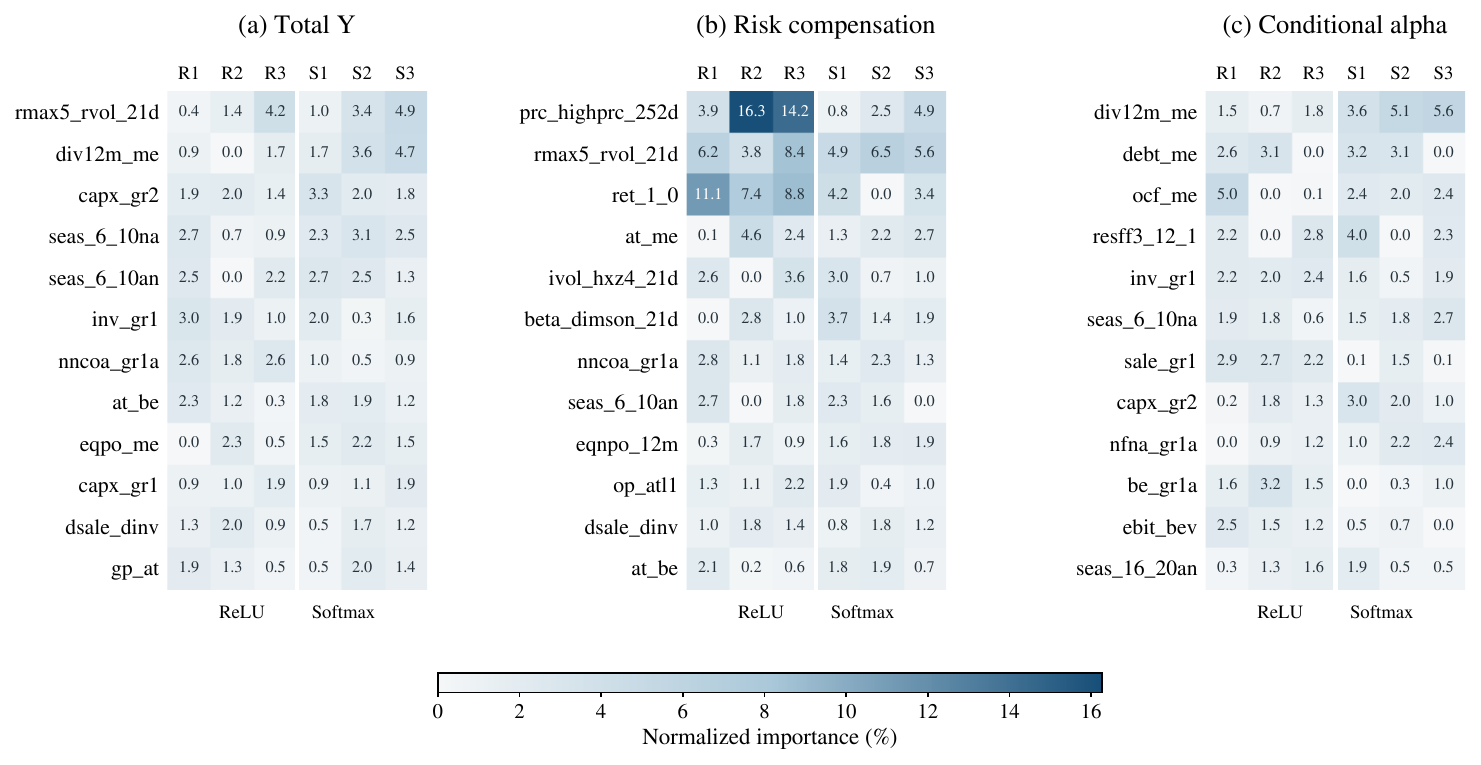}
    \caption{Leading characteristics of total return prediction and its components}
    \label{fig:rd_consensus_importance}
    \begin{minipage}{\textwidth}\small
    \textit{Notes:} Cells report normalized importance (\%). R1--R3 and S1--S3 denote ReLU and Softmax with one to three layers. Rows are ordered by average importance across architectures. All panels use a common color scale.
    \end{minipage}
    \end{figure}

    Figure~\ref{fig:rd_consensus_importance} compares the characteristics most relevant to total return prediction and its risk and alpha components. For each prediction task, we select twelve characteristics, prioritizing those with consistently high importance across attention mechanisms and using average ranks to complete the selection. Displaying their importance separately across the six architectures reveals both the information shared by the predictors and the distinctive emphasis of each component.
    
    The risk component emphasizes market related characteristics, including proximity to the annual price high, the maximum-return-to-volatility measure, and recent returns. Meanwhile, the alpha component draws more heavily on dividend yield, leverage, cash-flow yield, and inventory growth, alongside residual momentum and seasonality. Total Y combines these sources, with leading characteristics spanning market behavior, valuation, and investment.

\begin{figure}[!htbp]
	\centering
	\includegraphics[width=\textwidth]{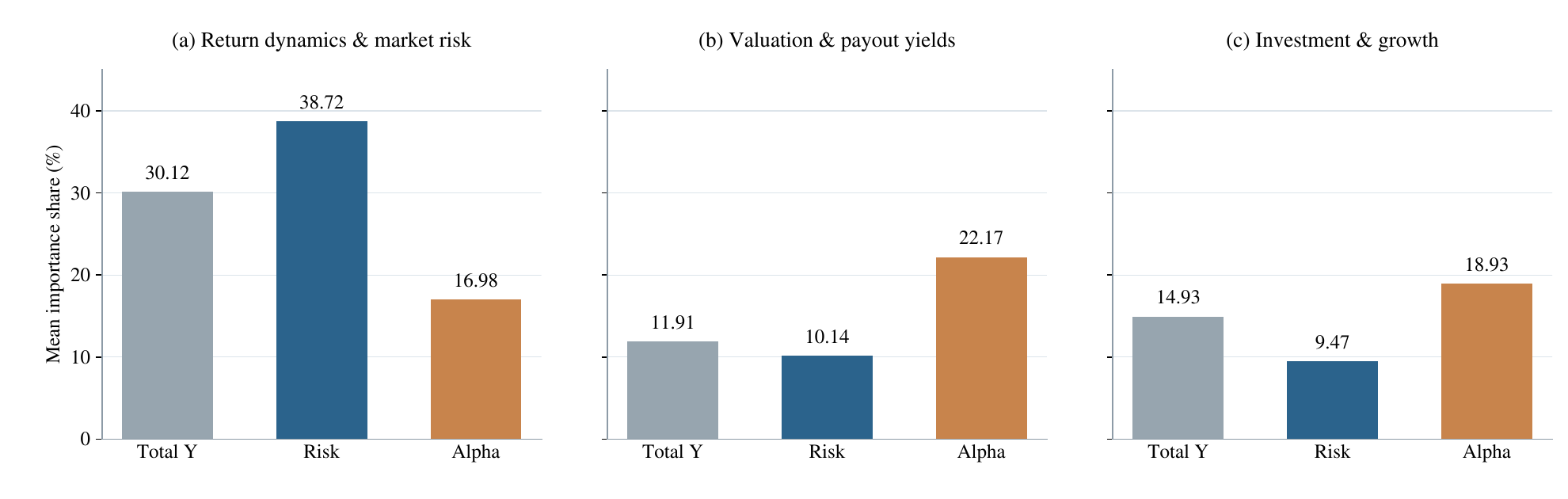}
	\caption{Economic composition of characteristic importance}
	\label{fig:rd_category_importance}
	\end{figure}

	Using the characteristic definitions documented by \citet{JensenKellyPedersen2023}, we organize selected characteristics into three broad economic groups: return dynamics and market risk (28 characteristics), valuation and payout yields (15), and investment and growth (22). Figure~\ref{fig:rd_category_importance} reports equal-weight averages across the six CAP-T architectures. Group shares sum constituent importance scores, retaining normalization over all 150 characteristics.

	Market-related characteristics account for 38.72\% of risk importance, compared with 30.12\% for total Y and 16.98\% for alpha. Valuation and payout yields instead receive their largest average share in alpha (22.17\%), followed by total Y (11.91\%) and risk (10.14\%). Investment and growth show a similar pattern, accounting for 18.93\% of alpha importance, 14.93\% of total Y importance, and 9.47\% of risk importance. Together, these patterns indicate that the estimated risk component places greater relative weight on market-related characteristics, whereas alpha emphasizes valuation and investment-related information. The latter pattern is consistent with evidence linking valuation and corporate investment to abnormal returns \citep{LakonishokShleiferVishny1994,TitmanWeiXie2004}. Total return prediction draws on both sets of characteristics, complementing the broader evidence on the predictive importance of market-based signals in \citet{GuKellyXiu2020}.

    \subsubsection{Testing conditional alpha}
    \label{sec:rd_alpha_persistence}
    
    The characteristic-importance results reveal distinct predictive roles for risk compensation and conditional alpha. We next examine whether the estimated alpha has a nonzero time average and whether its cross-sectional structure is preserved in out-of-sample forecasts. Using five factors and six-month updates, we study 167 firms from November 2009 to October 2020. Related to the pricing-error diagnostics of \citet{GuKellyXiu2021}, we focus on the time-average alpha.
    
    \begin{figure}[!htbp]
    \centering
    \includegraphics[width=\textwidth]{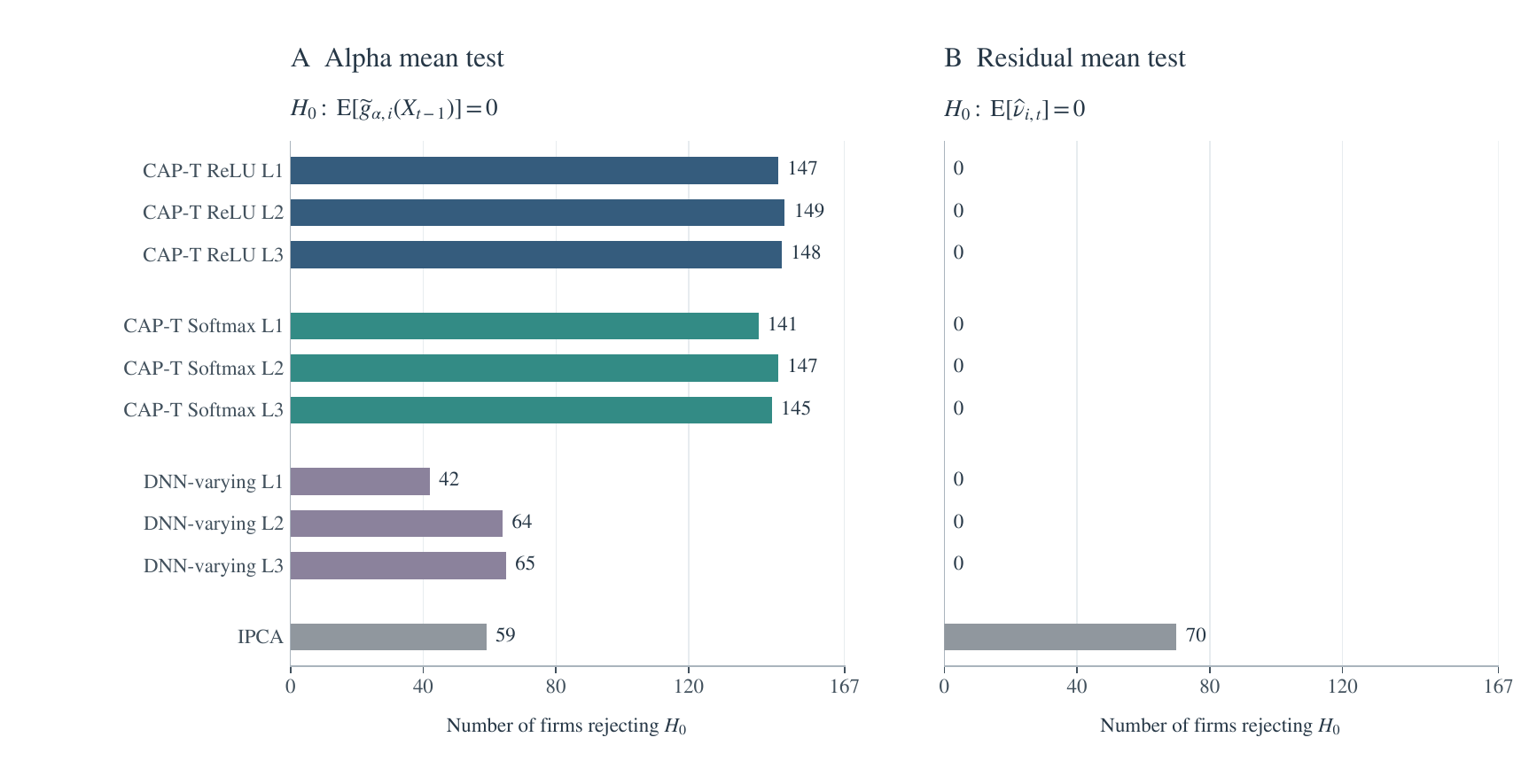}
    \caption{Tests of zero mean alpha and residuals}
    \label{fig:rd_alpha_mean_diagnostics}
    \begin{minipage}{\textwidth}\footnotesize
    \textit{Notes:} Horizontal bars show the number of firms rejecting a zero time mean among 167 firms. Panel A tests the mean of the decomposition estimate $\widetilde{g}_{\alpha,i}(X_{t-1})$; Panel B tests the mean of the fitted residual $\widehat{\nu}_{i,t}$ defined in Step S1. For CAP-T, the component identity in \eqref{eq:target_identities} gives the same residual after subtracting both decomposition components. Each model uses its own first-entry decomposition. Two-sided tests use Bartlett HAC standard errors with six lags \citep{NeweyWest1987} and 5\% BH correction within each model and panel \citep{BenjaminiHochberg1995}. Tests condition on fitted estimates.
    \end{minipage}
    \end{figure}
    
    Figure~\ref{fig:rd_alpha_mean_diagnostics} shows that the estimates identify a substantial average-return component
    outside the fitted risk-loading space. Across the six CAP-T architectures, 141–149 of the 167 firms exhibit nonzero average alpha after multiple-testing adjustment. This broad incidence persists across both attention mechanisms and all three network depths. The corresponding counts are 42–65 for DNN-varying and 59 for IPCA. Within the fitted CAP-T system, the alpha component therefore plays a substantive role in the return decomposition, and its presence is not confined to a particular architectural choice. The fitted residuals $\widehat{\nu}_{i,t}$ provide a complementary diagnostic. Across all six CAP-T architectures, no firm-level residual mean is significant after multiple-testing adjustment.

\begin{figure}[!htbp]
    	\centering
    	\includegraphics[width=\textwidth]{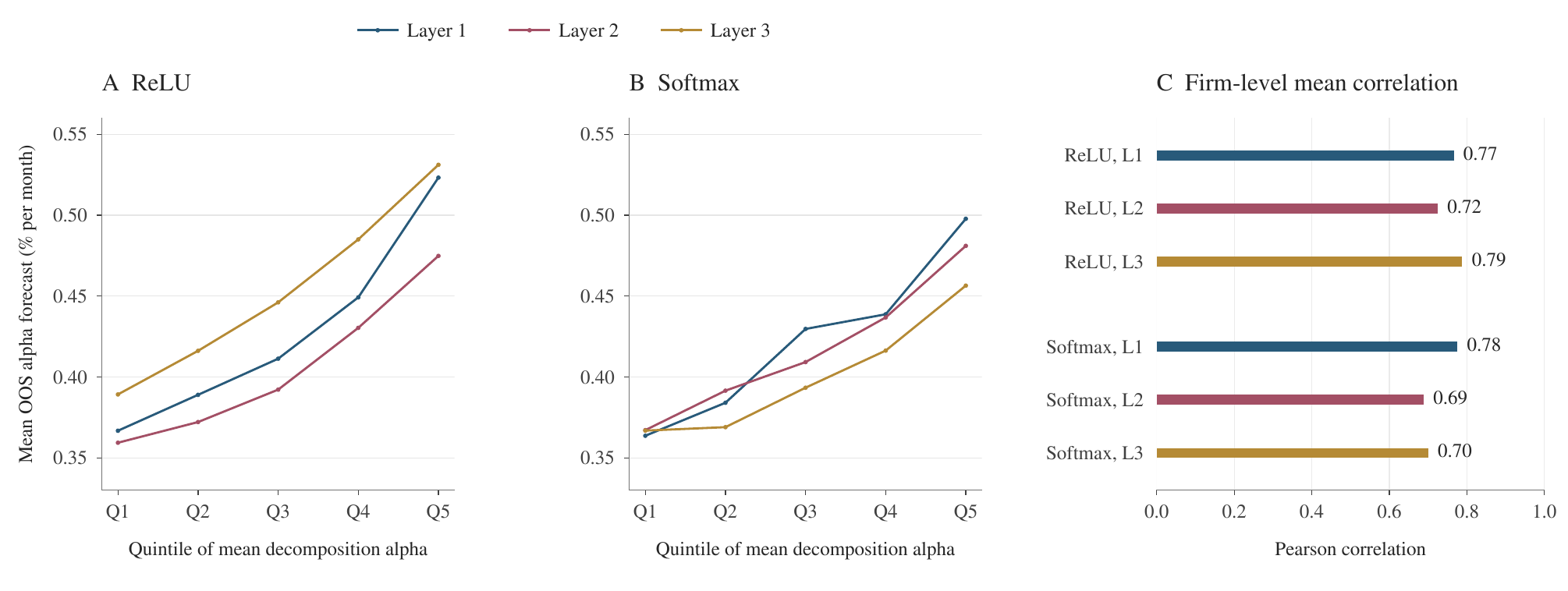}
    	\caption{Cross-sectional alpha structure in out-of-sample forecasts}
    	\label{fig:rd_alpha_forecast_structure}
    \end{figure}

    The preceding evidence identifies a persistent alpha component within the fitted return decomposition. Since portfolio formation uses predicted alpha to select stocks, we next examine whether out-of-sample alpha forecasts preserve its ordering across firms. For each architecture, we average fitted members within each month and calculate firm-level means over the same 132 months. Panels A and B of Figure~\ref{fig:rd_alpha_forecast_structure} group the 167 firms into quintiles of mean decomposition alpha. Mean forecast alpha increases monotonically from Q1 to Q5 under both attention mechanisms and all three network depths. Panel C reports correlations between the two firm-level means, ranging from 0.69 to 0.79. These results indicate that the forecasts retain cross-sectional differences relevant to stock selection.
    
	\section{Conclusion} \label{sec:conclusion}
	
	This paper develops CAP-T by combining cross-firm attention with an identified decomposition of conditional expected returns into alpha and risk compensation. The theoretical analysis establishes approximation and statistical error bounds under temporal dependence and traces error propagation through factor estimation and component learning. The empirical results document improved return prediction and useful portfolio signals after transaction costs. Component-level evidence reveals distinct patterns of characteristic importance and shows that out-of-sample alpha forecasts preserve the cross-sectional structure of decomposition alpha. Together, these findings demonstrate the value of combining flexible prediction with an economically interpretable return decomposition.
	
	However, the current implementation relies on structured firm characteristics. A natural extension is to enrich this information set with financial text and images, motivated in part by econometric methods for measuring news sentiment \citep{ShapiroSudhofWilson2022}. Integrating these sources within cross-firm attention would allow us to examine whether information beyond conventional characteristics primarily refines estimated risk compensation or captures additional conditional alpha. Because this extension introduces estimated representations alongside the estimated factor space, it also calls for inference that accounts for uncertainty across learning stages under temporal dependence. Indeed, the literature on multistep estimation provides a methodological starting point for this task \citep{ChenLiao2015}.

	\section*{Declaration of generative AI}
    The authors used OpenAI's ChatGPT and Codex to assist with language editing, code organization, and locating documentary evidence for the table of firm associations in the empirical analysis. The authors reviewed and verified the resulting text, code, and documentary evidence and take full responsibility for the final content.

	\printbibliography[heading=bibintoc,title={References}]

\end{document}